\documentclass{article}

\usepackage{arxiv}

\usepackage{multirow}
\usepackage{graphicx}
\usepackage[table]{xcolor}
\usepackage{multirow}
\usepackage{graphicx}
\usepackage{subcaption}
\usepackage{amsmath}
\usepackage{algorithm} 
\usepackage{algpseudocode}
\usepackage{enumitem}
\usepackage[page,header]{appendix}
\usepackage{titletoc}
\usepackage{wrapfig}
\usepackage{float}

\usepackage[utf8]{inputenc} % allow utf-8 input
\usepackage[T1]{fontenc}    % use 8-bit T1 fonts
\usepackage{xcolor}
\usepackage{hyperref}
\definecolor{citecol}{RGB}{192,79,21}
\definecolor{linkcol}{RGB}{48,105,170}
\definecolor{urlcol}{RGB}{48,105,170}
\hypersetup{ 
  colorlinks=true,
  urlcolor=urlcol,
  linkcolor=linkcol,
  citecolor=citecol,
}
\usepackage{url}            % simple URL typesetting
\usepackage{booktabs}       % professional-quality tables
\usepackage{amsfonts}       % blackboard math symbols
\usepackage{nicefrac}       % compact symbols for 1/2, etc.
\usepackage{microtype}      % microtypography
\usepackage[capitalize]{cleveref}       % smart cross-referencing
\usepackage{lipsum}         % Can be removed after putting your text content
\usepackage{graphicx}
\usepackage[numbers]{natbib}
\usepackage{doi}

\usepackage{pifont}
\newcommand{\cmark}{\textcolor{green!55!black}{\ding{51}}}
\newcommand{\xmark}{\textcolor{red!60!black}{\ding{55}}}

\title{Delving into the Temporal Challenges of Unified Video Protection Against Image-to-Video and Fine-Tuning-based Customization}

\date{}

\author{Yuxin Huang$^{1}$\quad Ziming Hong$^{1}$\quad Mingming Gong$^{2,5}$\quad Wanyu Wang$^{3}$\quad Jing Zhang$^{4}$\quad Tongliang Liu$^{1,5}$\\
$^{1}$Sydney AI Centre, The University of Sydney\quad  $^{2}$University of Melbourne \\
$^{3}$City University of Hong Kong\quad $^{4}$Wuhan University\\
$^{5}$Mohamed bin Zayed University of Artificial Intelligence
}

\renewcommand{\shorttitle}{}

\begin{document}
\maketitle

\begin{figure}[h!]
    \vspace{-14mm}
    \small
    \centering
    \includegraphics[width=0.99\textwidth]{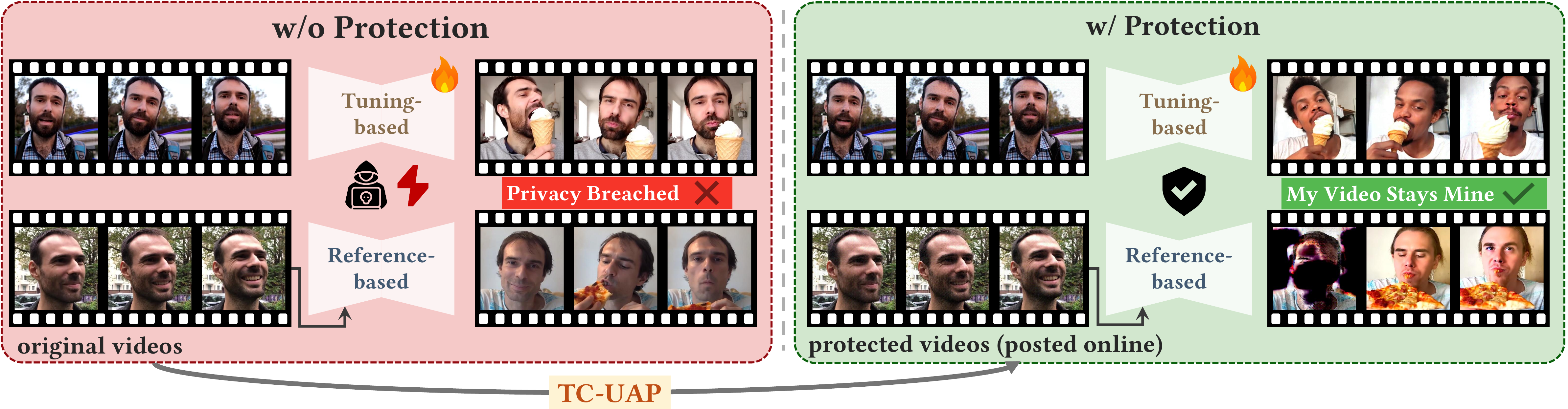}
    \caption{
    Protecting personal videos against unauthorized video customization. Once posted online, a user’s videos may be collected and exploited by either tuning-based or reference-based customization. Our \texttt{TC-UAP} proactively protects the videos by applying imperceptible temporally consistent perturbations, making both customization methods fail to generate usable results. The code, evaluation suite and additional results will be available on our \href{https://saythe17.github.io/TC-UAP/}{project page}.} 
    \label{fig:teaser}
\end{figure}

\begin{abstract}
Recent diffusion-based video generation models have enabled high-quality personalized video customization through both \textit{tuning-based pipelines}, which fine-tune a video diffusion model, and \textit{reference-based pipelines} such as image-to-video generation. However, these capabilities raise serious concerns about personal privacy, identity ownership and intellectual property protection. Existing anti-customization works focus on protecting images, while protection for videos against both reference- and tuning-based customization remains largely underexplored. Protecting videos in this setting raises three challenges: (i) Image-level perturbations, optimized frame by frame, cannot survive temporal compression by 3D video VAE. (ii) A video-level perturbation optimized on a single video is vulnerable to temporal editing and fails to protect unseen videos. (iii) Temporally inconsistent perturbations are not robust to temporal attacks. To address these challenges, we propose Temporally Consistent Universal Adversarial Perturbations (\texttt{TC-UAP}), the first protection method against both reference- and tuning-based video customization. 
\texttt{TC-UAP} optimizes an identity-level multi-frame UAP over sliding windows from multiple videos, accounting for local temporal dependencies induced by temporal compression in video VAE and enabling a single perturbation to protect unseen videos of varying lengths.
Moreover, we introduce intrinsic temporal modeling and an extrinsic surrogate temporal-attack loss, which make the perturbation temporally consistent and robust to unseen temporal attacks. Empirically, quantitative and qualitative results show that \texttt{TC-UAP} achieves the strongest identity protection compared with existing methods under both reference- and tuning-based video customization, and remains robust under multiple unseen temporal attacks.
\end{abstract}

\vspace{-1mm}
\section{Introduction}
\label{sec:introduction}
\vspace{-1mm}

Recent advances in diffusion models have revolutionized video creation, enabling the generation of high-quality videos. Both open-source~\cite{kong2024hunyuanvideo, yang2024cogvideox, wan2025wanopenadvancedlargescale, HaCohen2024LTXVideo} and commercial models~\cite{seedance2026seedance, zheng2025open} have shown increasingly impressive results. 
These models enable users to specify the subject of the generated video, a capability commonly known as video customization. Such customization typically follows two approaches. \textit{Tuning-based methods}~\cite{abdal2025dynamic} fine-tune a video diffusion model on a small set of video clips, for example, using Low-Rank Adaptation (LoRA)~\cite{hu2022lora}, and can then generate diverse new videos. \textit{Reference-based methods}~\cite{chen2026identity, guo2026dreamid, he2024id, yuan2025identity}, such as image-to-video (I2V) models, take a reference image and a text prompt as input, offering pixel-level control.

However, these models raise serious concerns about personal privacy, identity ownership and intellectual property protection \cite{juefei2022countering,li2025rethinking,li2026webcloak,zhao2026intellectual,hong2025toward,feng2025token,wang2021non} in video content. From public video clips of a target person, a malicious user can exploit current models to generate misleading or harmful content \cite{miao2024t2vsafetybench,xiang2026safety,zheng2026vii,ma2026safegen,lin2026force} about that person through either approach: finetuning a video diffusion model on the whole set through tuning-based customization, or extracting a single frame from any of these videos as the reference image for reference-based customization. These threats may lead to fake news and violations of portrait rights. Protecting video content from being misused for unauthorized customization thus becomes a pressing concern.

Existing anti-customization works have mostly focused on adversarial perturbation-based image-level protection. Early efforts mainly protect \textit{images} against tuning-based customization of text-to-image (T2I) models~\cite{liang2023adversarial, shan2023glaze, van2023anti, liang2023mist, liu2024metacloak, wang2024simac, wu2025myopia, zhao2024unlearnable, teng2025idcloakcraftingidentityspecificcloaks, hu2025targeted} and reference-based image-to-image (I2I) editing~\cite{pmlr-v202-salman23a, bala2025dctshield, shao2026unidef, lee2026universal}. More recent works have further extended image protection to video generation scenarios, where the protected content is still a reference image used by image-to-video (I2V) pipelines~\cite{pang2024vgmshield, gui2025i2vguard, vu2026anti, chowdhury2025vid, long2026immune2v}. \textit{In contrast, protecting videos has received much less attention, despite the growing prevalence of video-based content across online platforms.} 
This limitation is particularly concerning because publicly available videos can be exploited as both \textit{references} or \textit{training data} for the two kinds of unauthorized customization. However, protecting videos against both reference- and tuning-based customization remains largely unexplored, highlighting the need for a unified video protection method against both types of threats.

To this end, we explore protecting videos from reference- and tuning-based customization, and identify three temporal challenges\footnote{The complete analysis and additional evidence for the three temporal challenges are provided in \cref{sec:rethink}.}, namely, (i) \textit{temporal compression}, (ii) \textit{temporal overfitting}, and (iii) \textit{temporal inconsistency}:
\begin{itemize}[leftmargin=*, topsep=0pt]\setlength{\parskip}{0pt}
    \item (i) \textit{Image-level perturbations, optimized frame by frame, cannot survive temporal compression in 3D video VAE.} 
    Image-level protection can be applied to videos by optimizing perturbations frame by frame. While such perturbations remain applicable to reference-based I2V customization, they are insufficient for tuning-based video customization, where a 3D causal video VAE encodes the entire video into compact latent representations~\cite{yang2024cogvideox, kong2024hunyuanvideo, wan2025wanopenadvancedlargescale, HaCohen2024LTXVideo}. 
    This \textit{temporal compression} substantially weakens frame-wise perturbations, causing them to fail after video encoding, as illustrated in \cref{fig:obs1a}. Therefore, protecting videos against both threats requires a video-level perturbation design.
    \item (ii) \textit{A perturbation optimized on a single video is vulnerable to temporal editing and fails to generalize to unseen videos.} 
    Optimizing a perturbation for a single video tends to overfit to the original temporal sequence. When the protected video undergoes temporal editing (e.g., trimming into a shorter clip), the protection may significantly degrade, as shown in \cref{fig:obs2}. Moreover, such a video-specific perturbation cannot be directly transferred to unseen videos. These limitations pose challenges to temporal flexibility, identity-level generalization, and protection efficiency.
    \item (iii) \textit{Temporally inconsistent perturbations are not robust to malicious temporal attacks.} 
    An attacker can exploit cross-frame redundancy across video frames to perform temporal attacks (e.g., frame interpolation and temporal averaging), which can substantially attenuate temporally inconsistent perturbations, as shown in \cref{fig:obs3}. This highlights the necessity of modeling temporal consistency of perturbations to ensure robust protection against temporal attacks.
\end{itemize}

To address these challenges, we propose \underline{T}emporally \underline{C}onsistent \underline{U}niversal \underline{A}dversarial \underline{P}erturbations (dubbed \texttt{TC-UAP}), the first video protection method against both reference- and tuning-based video customization, as shown in \cref{fig:teaser}. 
Specifically, we model the protection signal as an identity-level multi-frame UAP comprising multiple temporally indexed perturbation frames, thereby providing greater temporal expressiveness for disrupting the video VAE latent representations shared by both customization pipelines. Such a multi-frame UAP can be temporally repeated and applied to videos of different lengths, enabling it to be reused across clips without per-video optimization.
Moreover, to make the perturbation survive the \textit{temporal compression} of 3D video VAEs while avoiding \textit{temporal overfitting}, we adopt a \textit{sliding-window optimization} strategy over multiple videos of the same identity. The window length is chosen to cover the dominant portion of the temporal receptive field, allowing each window to capture sufficient temporal dependencies induced by temporal compression. By sampling windows from different videos and temporal positions, this strategy exposes the UAP to diverse temporal contexts and prevents it from becoming tied to the complete temporal sequence of any training video. In addition, although the multi-frame formulation improves temporal expressiveness, it does not by itself guarantee temporal consistency. We therefore introduce \textit{intrinsic temporal modeling} and an \textit{extrinsic surrogate temporal-attack loss} to regularize cross-frame consistency and preserve protection effectiveness under temporal attacks. 
As a result, once optimized, the same UAP can be efficiently reused to protect unseen videos of the same identity while remaining robust against unseen temporal attacks.

Empirically, we conduct extensive experiments to evaluate \texttt{TC-UAP}. Both the quantitative and qualitative results show that our method achieves the strongest identity protection compared with existing methods under both reference- and tuning-based video customization, and remains robust under diverse unseen temporal attacks. Our contributions are summarized as follows:
\begin{itemize}[leftmargin=*, topsep=0pt]\setlength{\parskip}{0pt}
    \item We are the first to formulate and study the problem of unified video protection against both reference- and tuning-based customization. We identify three key temporal challenges and provide empirical evidence of their impact on protection effectiveness: (i) \textit{temporal compression}, (ii) \textit{temporal overfitting}, and (iii) \textit{temporal inconsistency}.
    
    \item We propose \texttt{TC-UAP}, a unified video protection method that models the protection signal as an identity-level multi-frame UAP, thereby enhancing temporal expressiveness and enabling efficient reuse across videos of different lengths. We adopt a sliding-window optimization strategy to mitigate temporal overfitting, while intrinsic temporal modeling and a surrogate temporal-attack loss improve robustness against unseen temporal attacks.
    \item Extensive quantitative and qualitative experiments demonstrate that \texttt{TC-UAP} provides effective identity protection against both reference- and tuning-based video customization, generalizes to unseen videos of the same identity, and remains robust against a range of unseen temporal attacks.
    
    \item We also establish and release a unified evaluation suite for video protection against both reference- and tuning-based customization, covering imperceptibility, protection effectiveness, transferability, and robustness. The suite provides standardized evaluation protocols and implementations for reproducible comparison of future methods.
\end{itemize}

\section{Related Work}
\label{sec:related-work}

\textbf{Image protection against generative models. }
Existing image protection methods mainly aim to protect images from two types of unauthorized use. 
For \textit{tuning-based customization}, where malicious attackers collect multiple images to fine-tune a text-to-image model via DreamBooth~\cite{ruiz2023dreambooth} or LoRA~\cite{hu2022lora}, existing defenses typically add imperceptible adversarial perturbations to the images, making the fine-tuned model fail to capture the identity or artistic style~\cite{liang2023adversarial, shan2023glaze, liang2023mist, van2023anti, liu2024metacloak, wang2024simac, wu2025myopia, zhao2024unlearnable, teng2025idcloakcraftingidentityspecificcloaks}. 
For \textit{reference-based customization}, existing defenses mainly protect images which can be used as reference images for image-to-image (I2I) editing~\cite{brooks2023instructpix2pix, wan2026mobile, wan2025mft} and image-to-video (I2V) generation. Specifically, imperceptible adversarial perturbations are added to the reference image, leading to distorted editing results in I2I editing~\cite{pmlr-v202-salman23a, wang2025ace, bala2025dctshield, lee2026universal, shao2026unidef, song2025idprotector, hong2025adlift} and low-quality videos or unstable motion in I2V generation~\cite{pang2024vgmshield, gui2025i2vguard, vu2026anti, chowdhury2025vid, long2026immune2v}. Recent work further introduces a benchmark for evaluating image protection methods in I2V generation~\cite{li2026ipbench}. However, existing works focus on protecting images, leaving video-level protection against both reference- and tuning-based customization unexplored.

\textbf{Video protection against unauthorized usage. }
Recent works have explored video protection under different unauthorized scenarios, including object tracking~\cite{wu2025temporal}, malicious video editing~\cite{li2024prime,cao2025videoguard}, and style mimicry on video imagery~\cite{passananti2024disrupting}. However, these video protection methods do not address unauthorized personalized customization using video diffusion models. Notably, although Gimbal~\cite{passananti2024disrupting} targets video content, it still performs image-level protection by protecting extracted video frames from being used to fine-tune text-to-image (T2I) diffusion models.

\textbf{Customizable video generation models. }  Recent video generation models~\cite{kong2024hunyuanvideo, yang2024cogvideox, wan2025wanopenadvancedlargescale, HaCohen2024LTXVideo, wang2025lavin, wang2025taming, zheng2026aligning} enable the same two types of customization discussed in image-protection works above. For \textit{tuning-based customization}, open-source models such as LTX-Video~\cite{HaCohen2024LTXVideo}, Wan~\cite{wan2025wanopenadvancedlargescale}, CogVideoX~\cite{yang2024cogvideox}, and HunyuanVideo~\cite{kong2024hunyuanvideo} allow users to train LoRA adapters~\cite{hu2022lora} on a few videos and then generate new videos from text prompts. For \textit{reference-based customization}, a pretrained image-to-video (I2V) model is conditioned on a single reference image. Open I2V pipelines include LTX-2.3 I2V~\cite{hacohen2026ltx2efficientjointaudiovisual}, Wan2.2-5B TI2V~\cite{wan2025wanopenadvancedlargescale}, CogVideoX-5b-I2V~\cite{yang2024cogvideox}, Stable Video Diffusion~\cite{blattmann2023stable}. 
Together, these tools make video customization broadly accessible and raise concerns about identity misuse, highlighting the need for video protection against diffusion-based customization.

\section{Preliminaries}
\label{sec:preliminaries}

\subsection{Problem Formulation}
\label{subsec:problem-formulation}

Consider a set of $M$ videos $\mathcal{V}_y=\{V_i\}_{i=1}^{M}$ of an identity $y$.
Each video is denoted as $V_i=\{v_i^t\}_{t=0}^{T_i-1}\in\mathbb{R}^{T_i\times C\times H\times W}$, where $T_i$ is the number of frames, and $C$, $H$, and $W$ denote the channel, height, and width of each frame $v_i^t$.

An attacker may use these videos for either \textit{tuning-based} or \textit{reference-based} customization, which we introduce in \cref{subsec:video-customization}. 
In both cases, the attacker aims to synthesize new videos that reproduce the target identity $y$.

The \textit{goal of the protector} is to construct protected videos that remain nearly indistinguishable from the originals, while preventing both reference- and tuning-based customization from generating usable videos that faithfully preserve the target identity $y$.
Specifically, successful protection should induce customization failure in the generated videos, reflected by \textit{degraded identity preservation, weakened prompt alignment, and reduced visual quality}.

Following the common perturbation-based protection paradigm \cite{liang2023adversarial, shan2023glaze, van2023anti, pang2024vgmshield, gui2025i2vguard, vu2026anti, chowdhury2025vid, long2026immune2v}, the protection goal can be formulated as adding imperceptible perturbations $\delta = \{\delta_i\}_{i=1}^M$ to $\mathcal{V}_y$ to obtain a protected video set $\mathcal{V}_y'=\{V_i'\}_{i=1}^{M}$.
For each video $V_i$, the protector generates a perturbation $\delta_i$ and obtains:
\begin{equation}
V_i' = V_i + \delta_i, \quad \text{s.t.} \quad ||\delta_i||_\infty \le \eta,
\end{equation}
where $\eta$ denotes the perturbation budget. The perturbation optimization objective is defined as:
\begin{equation}
\delta^{\star}
=
\arg\min_{||\delta||_\infty \le \eta}
\mathcal{L}_{\text{id}}(\mathcal{V}_y, \delta),
\label{eq:protect-obj}
\end{equation}
where $\mathcal{L}_{\text{id}}$ is a surrogate objective, and minimizing it induces the customization failure defined above.

\subsection{Video Customization}
\label{subsec:video-customization}
\textbf{Diffusion Models. }
Diffusion Models (DMs) are generative models~\cite{song2020denoising, ho2020denoising, Lin_2026_CVPR, lin2026subflow} that learn a target data distribution. Given a clean sample $z_0$, the forward process produces a noisy sample at step $\tau$ as $z_\tau = \sqrt{\bar{\alpha}_\tau}\, z_0 + \sqrt{1 - \bar{\alpha}_\tau}\, \epsilon, \quad \epsilon \sim \mathcal{N}(0, I)$, where $\{\bar{\alpha}_\tau\}$ is a fixed noise schedule. A denoising network $\epsilon_\omega$ is trained to predict $\epsilon$ from $z_\tau$ under a conditioning signal $c$ (e.g., text, image, etc.) encoded by a corresponding encoder $\mathcal{E}_{\text{text}}$:
\begin{equation}
    \mathcal{L}_{\text{DM}}(\omega) = \mathbb{E}_{z_0,\, \tau,\, \epsilon,\, c} \big[\, \big\| \epsilon - \epsilon_\omega(z_\tau,\, \tau,\, \mathcal{E}_{\text{text}}(c)) \big\|_2^2 \,\big].
    \label{eq:ldm}
\end{equation}

\textbf{Video VAE. }
Latent diffusion models (LDMs)~\cite{rombach2022high} operate in a compressed latent space encoded by an encoder $\mathcal{E}$. For video diffusion, it is instantiated as a video VAE that operates along the spatial and temporal dimensions. Given a video $V_i \in \mathbb{R}^{T_i \times C \times H \times W}$, the encoder produces a latent:
\begin{equation}
    z_i = \mathcal{E}(V_i) \in \mathbb{R}^{\tilde{V}_i \times \tilde{C} \times \tilde{H} \times \tilde{W}}, \qquad \tilde{V}_i = \tfrac{T_i-1}{r_t} + 1,\ \ \tilde{H} = \tfrac{H}{r_h},\ \ \tilde{W} = \tfrac{W}{r_w},
    \label{eq:vae}
\end{equation}
which is downsampled by factors $r_h$ and $r_w$ along the spatial dimension and $r_t$ along the temporal dimension, with $\tilde{C}$ latent channels. We use $t\in\{0,\dots,T_i-1\}$ as the frame index in the input video, and $p\in\{0,\dots,\tilde{V}_i-1\}$ as the temporal position in the latent space. The decoder $\mathcal{D}$ reconstructs $\hat{V}_i = \mathcal{D}(z_i)$.

\textbf{Tuning-based Video Customization. }
Tuning-based methods fine-tune a pretrained denoising network $\epsilon_\omega$ on a set of video-text pairs $\{(V_i, c_i)\}$ with the latents $z_i = \mathcal{E}(V_i)$. A lightweight adapter (e.g., LoRA~\cite{hu2022lora}) parameterized by $\Delta\omega$ is optimized with the denoising objective:
\begin{equation}
    \mathcal{L}_{\text{tune}}(\Delta\omega) = \mathbb{E}_{(V_i, c_i) \sim \{(V_i, c_i)\},\, \tau,\, \epsilon} \big[\, \big\| \epsilon - \epsilon_{\omega + \Delta\omega}(z_{i,\tau},\, \tau,\, \mathcal{E}_{\text{text}}(c_i)) \big\|_2^2 \,\big],
    \label{eq:tune}
\end{equation}
where $z_{i,\tau}$ is the noisy latent at step $\tau$. Given the adapted model $\epsilon_{\omega + \Delta\omega}$ and a new prompt $c^{\star}$, a new video is generated by iterative denoising in the latent space, followed by VAE decoding.

\textbf{Reference-based Video Customization. }
Reference-based methods take a reference image $I_{\text{ref}}$ and a prompt $c$, and generate a video through a pretrained image-to-video (I2V) model $g_\phi$. The reference is first encoded as $z_{\text{ref}} = \mathcal{E}(I_{\text{ref}}) $ and injected into the denoising trajectory in one of two common ways:
\begin{equation}
    \epsilon_\phi\!\left(z_\tau,\, \tau,\, \mathcal{E}_{\text{text}}(c),\, \mathcal{A}(z_{\text{ref}})\right) \quad\text{or}\quad \epsilon_\phi\!\left([\,z_\tau \,\Vert\, z_{\text{ref}}\,],\, \tau,\, \mathcal{E}_{\text{text}}(c)\right),
    \label{eq:ref-inject}
\end{equation}
where $\mathcal{A}(\cdot)$ denotes a cross-attention branch and $[\,\cdot \,\Vert\, \cdot\,]$ denotes channel-wise concatenation. The denoised latent is finally decoded back to a video via VAE decoding.

\section{Rethinking Video Protection}
\label{sec:rethink}

In this section, we rethink video protection from the perspective of video temporal structure. We first consider a general protection objective against both reference- and tuning-based customization, and then show that naively transferring image protection methods to videos leads to several failure modes. These observations reveal three video-specific temporal challenges that motivate our method design.

\subsection{Video VAE as the Shared Vulnerability}

Diffusion-based video customization typically starts by encoding videos or images into latent representations using a 3D video VAE, which compresses high-dimensional data (e.g., raw pixels) into compact latents for computational efficiency. Reference-based methods~\cite{hacohen2026ltx2efficientjointaudiovisual, wan2025wanopenadvancedlargescale, blattmann2023stable} encode the reference image before generation, while tuning-based methods~\cite{abdal2025dynamic, hu2022lora} encode training videos before fine-tuning. This makes the VAE latent a shared representation in both customization pipelines. We thus optimize the protective perturbation to disrupt the VAE latent, so that identity information is corrupted before being used by downstream customization. 

To analyze the protection effectiveness and identify video-specific challenges, we use the \textbf{VAE reconstruction} result in this section. An accurate reconstruction indicates that the latent still contains usable information, and a corrupted one indicates that this information has been disrupted. All experiments in this section are conducted using the 3D causal video VAEs of LTX-2.3~\cite{hacohen2026ltx2efficientjointaudiovisual} and Wan2.2~\cite{wan2025wanopenadvancedlargescale}.

\begin{figure}[t]
    \centering
    \includegraphics[width=\textwidth]{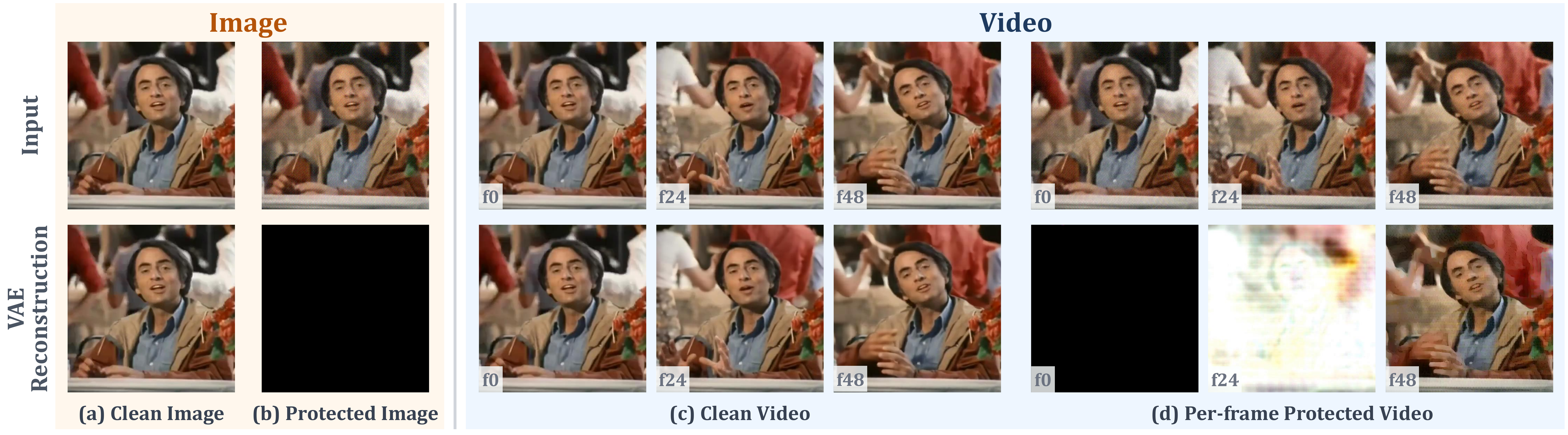}
    \caption{Image-level protection under video VAE reconstruction. (a) Clean image. (b) Protected image. (c) Clean video, \textit{where the numbers indicate frame indices}. (d) Protected video obtained by applying image protection to each frame.
    Images are treated as single-frame videos, and all inputs are reconstructed by the same video VAE. The top row shows the inputs to the video VAE, and the bottom row shows the corresponding VAE reconstructions.}
    \label{fig:obs1a}
\end{figure}

\begin{figure}[t]
    \centering
    \includegraphics[width=\textwidth]{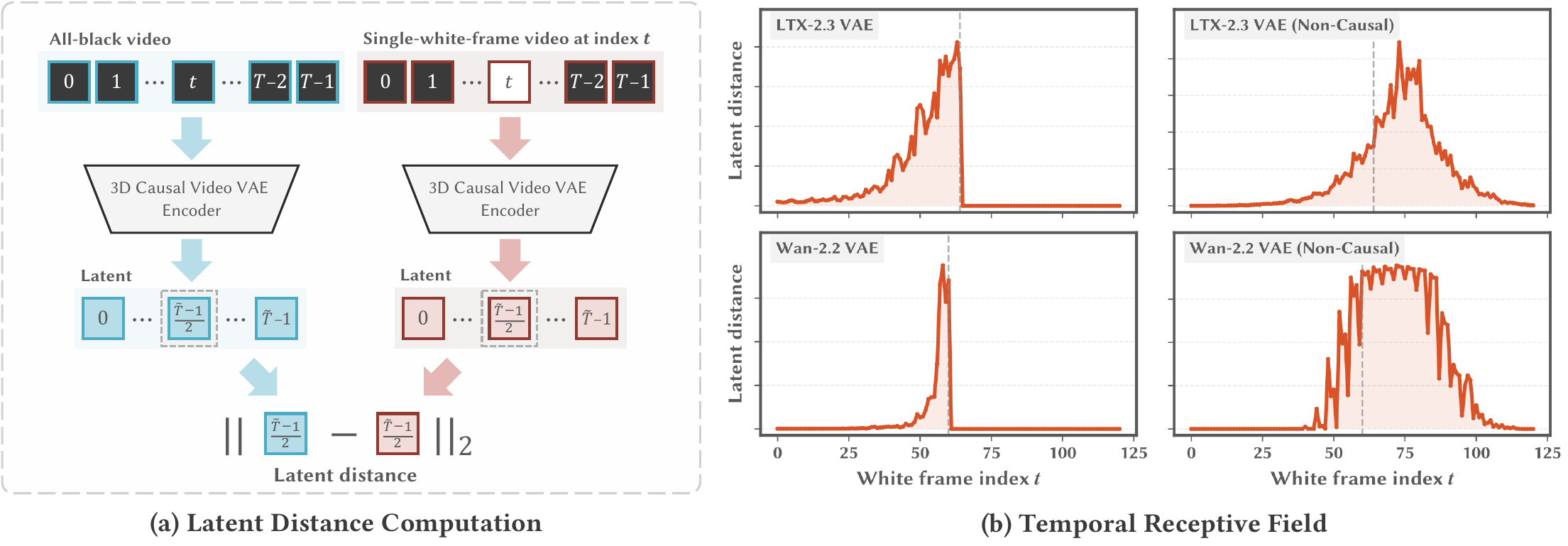}
    \caption{Temporal receptive-field analysis of 3D causal video VAEs.     
    (a) Latent distance computation. We encode an all-black video and a single-white-frame video, in which only the $t$-th frame is changed to white, and compute their \textit{latent distance at the middle temporal position}. The middle latent position is defined as the rounded-down midpoint of the latent temporal positions.
    (b) Temporal receptive fields of LTX-2.3~\cite{hacohen2026ltx2efficientjointaudiovisual} and Wan-2.2~\cite{wan2025wanopenadvancedlargescale} VAEs. The x-axis denotes the white-frame index $t$, and the y-axis shows the corresponding latent distance computed in (a). The left and right columns show the original causal and modified non-causal versions. Larger latent distance indicates stronger influence.}
    \label{fig:obs1b}
\end{figure}

\subsection{Temporal Challenges in Video Protection}
\label{sec:temporal-challenges}

Unlike existing image protection methods for T2I and I2I generation, and even I2V generation where protection is usually applied to a single image reference, protecting complete videos requires addressing additional challenges arising from temporal structure. We identify three key temporal challenges that hinder effective full-video protection: \textit{temporal compression}, \textit{temporal overfitting}, and \textit{temporal inconsistency}.

\textbf{Temporal compression: }\textbf{image-level perturbations cannot survive \textit{temporal compression} in 3D video VAE.} 
A straightforward way to protect videos is to extend image-level protection in a frame-wise manner, i.e., by optimizing an image-level perturbation against video-based customization independently for each frame. 
We first consider this frame-wise image-level setting.
Given a frame image $v$, an image-level perturbation $\xi$ can be optimized by maximizing the distance between the VAE latent representations of the clean and perturbed inputs:
\begin{equation}
\xi^{\star} = \arg\max_{\|\xi\|_\infty \le \eta}
\big\|\mathcal{E}(v+\xi) - \mathcal{E}(v)\big\|_2^2,
\label{eq:image-level-protection}
\end{equation}
where $\mathcal{E}$ is the encoder of the 3D causal video VAE. As shown in \cref{fig:obs1a}(a-b), the \textit{clean image} is accurately reconstructed, whereas the \textit{protected image} is reconstructed as a near-black image, indicating that single-image protection is effective. 
\textit{However, this effectiveness does not carry over to videos.} From the comparison between \cref{fig:obs1a}(c-d), when the same image-level objective is optimized and applied frame by frame for the video, although the reconstructed video is initially disrupted, the reconstruction \textit{gradually recovers} in later frames. 

We attribute this phenomenon to the \textbf{temporal compression} mechanism of 3D \textbf{causal} video VAEs: 
\begin{itemize}[leftmargin=*, topsep=0pt]\setlength{\parskip}{0pt}
    \item \textbf{Temporal compression weakens image-level perturbations.}
    The video VAE encoder compresses information from multiple frames into each temporal latent position. Since image-level perturbations are optimized independently for each frame, even if they effectively disrupt the corresponding single-frame latent representation, \textit{their effect can be weakened after being aggregated with temporally compressed information from neighboring frames}. In other words, a perturbation that enlarges the latent distance for an individual frame does not necessarily enlarge the distance between the temporally compressed latent representations of the clean and protected videos\footnote{This temporal compression is different from directly removing perturbations via pixel-space or spatial denoising. We verify this with two variants: (i) the static-video setting in \cref{fig:static}; and (ii) image-level Universal Adversarial Perturbations (UAPs) in \cref{fig:1f_uap}.}.

    \item \textbf{Causal temporal modeling makes the degradation more severe in later frames.}
    Due to causal temporal modeling, such as causal temporal convolution~\cite{wan2025wanopenadvancedlargescale,hacohen2026ltx2efficientjointaudiovisual}, the latent representation at each temporal position can only aggregate information from the current and previous frames, while future frames are inaccessible\footnote{We analyze the \textit{temporal receptive field} of the 3D causal video VAE. As illustrated in \cref{fig:obs1b}(a), we measure the influence of each frame index $t$ by comparing the middle-position latent of an all-black video and a single-white-frame video. \cref{fig:obs1b}(b) (left column) shows that earlier frames can still affect the middle temporal position, indicating that each latent is obtained through temporal compression and information aggregation over previous frames. The influence decays with temporal distance from the middle latent position, showing that temporally distant previous frames contribute less.}.
    As a result, later frames aggregate temporally compressed information from more previous frames, leading to stronger perturbation degradation. In contrast, earlier frames involve a shorter temporal history and are therefore less affected\footnote{Notably, if we close the causal mechanism in 3D video VAE, earlier frames also lose protection effectiveness (\cref{fig:obs1_app}). This is because the non-causal temporal receptive field also aggregates information from later frames, as shown in \cref{fig:obs1b}(b) (right column).}.

\end{itemize}

This suggests that effective video protection should optimize perturbations over multiple frames jointly, so as to account for temporal dependencies introduced by temporal compression and causal temporal modeling.

\begin{figure*}[t]
    \centering
    \includegraphics[width=\textwidth]{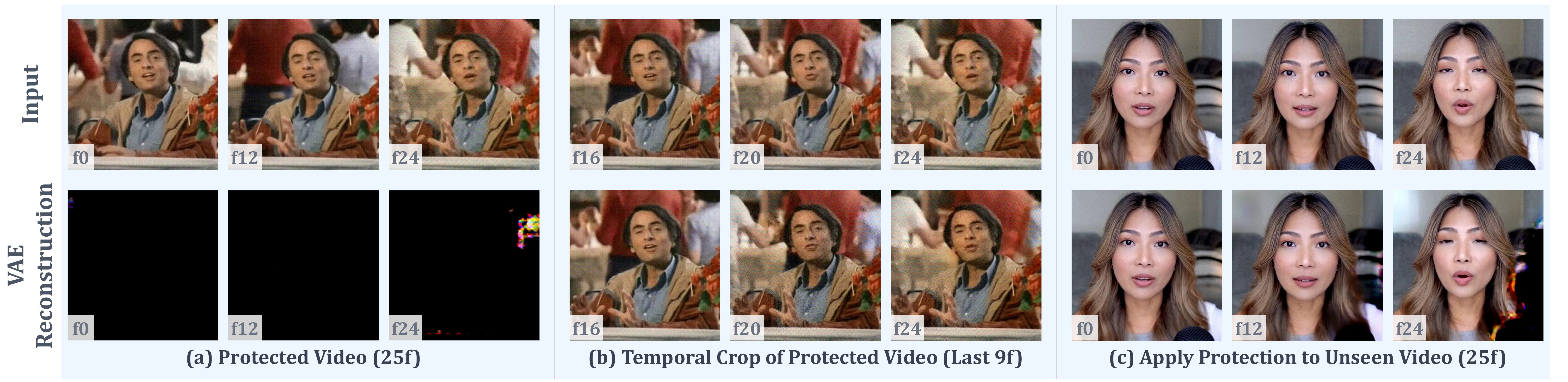}
    \caption{Per-video perturbations overfit to the optimized video. (a) A perturbation optimized on a 25-frame video provides strong protection for the original video. (b) When the protected video is temporally cropped to its last 9 frames, the protection largely disappears. (c) Applying the same perturbation to an unseen 25-frame video fails to protect it.}
    \label{fig:obs2}
\end{figure*}

\textbf{Temporal overfitting: }\textbf{a perturbation optimized on a single video is vulnerable to \textit{temporal editing} and \textit{fails to generalize} to unseen videos.} 
A natural next step is therefore to optimize perturbations at the video level. 
However, simply moving from image-level perturbations in a frame-wise manner to video-level perturbations for the full video is not enough. 
Specifically, a full video-level perturbation can easily become temporally inflexible: it may be vulnerable to \textit{temporal edits} such as trimming or length changes, fail to generalize across different clips, and incur optimization costs that scale with the full video length.
To examine this, we consider a straightforward video-level baseline that applies Projected Gradient Descent (PGD)~\cite{madry2017towards} by directly optimizing a video-level perturbation $\xi \in \mathbb{R}^{T \times C \times H \times W}$ with the same shape as the input video:
\begin{equation}
    \xi^{\star} \;=\; \arg\max_{\|\xi\|_\infty \le \eta}
    \big\|\,\mathcal{E}(V+\xi) - \mathcal{E}(V)\,\big\|_2^2.
    \label{eq:per-video-baseline}
\end{equation} 
As shown in \cref{fig:obs2}(a), this baseline provides strong protection across all frames, showing that video-level optimization can disrupt video VAE reconstruction. However, the optimized perturbation overfits to the specific temporal sequence of the video used for optimization, leading to poor generalizable protection in two aspects:
\begin{itemize}[leftmargin=*, topsep=0pt]\setlength{\parskip}{0pt}
    \item \textbf{Vulnerability to temporal editing of the same video.} When the protected video is temporally cropped (e.g., to its last 9 frames), the protection largely disappears, as shown in \cref{fig:obs2}(b). This failure indicates that optimizing a perturbation over the complete video can cause \textit{temporal overfitting} to the specific temporal sequence, making the protection sensitive to even common temporal edits such as changing the temporal span or starting position.
    \item \textbf{Failure to generalize to unseen videos.} Applying the same perturbation to an unseen 25-frame video also fails to disrupt the reconstruction, as shown in \cref{fig:obs2}(c). This indicates that per-video PGD is not reusable across different video content, and a new perturbation must be optimized for each video.
\end{itemize}
These results suggest that effective video protection should avoid overfitting to a specific temporal sequence, thereby remaining robust to temporal cropping or length changes. Ideally, it should also generalize to unseen clips, reducing the need to optimize a new perturbation for every full video and thus enabling practical deployment.

\begin{figure}[t]
    \centering
    \includegraphics[width=0.8\textwidth]{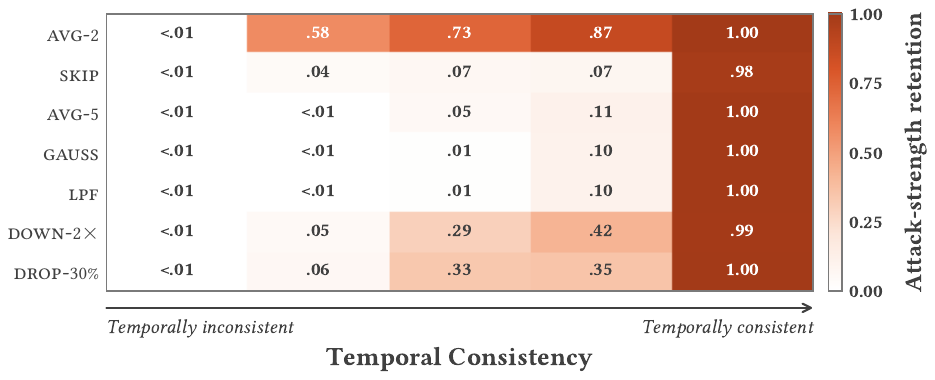}
    \caption{Attack-strength retention under 7 temporal attacks (named in the left column, formulations in Appendix~\ref{app:rethinking-2}). Temporal consistency is evaluated by cosine similarity between adjacent frames. }
    \label{fig:obs3}
\end{figure}

\textbf{Temporal inconsistency: }\textbf{temporally inconsistent perturbations are not robust to temporal attacks.} Before using a protected video for customization, an attacker can apply \textit{temporal attacks}, such as frame interpolation and frame averaging, \textit{to weaken the protection by attenuating the perturbation in pixel space} before it is encoded by the video VAE. We observe that when each frame of the video perturbation is optimized independently, the perturbation becomes highly vulnerable to such temporal attacks. We refer to such a perturbation as \textbf{temporally inconsistent}. To quantify how well a perturbation survives a temporal attack $\mathcal{T}$, we measure the attack-strength retention. As shown in \cref{fig:obs3}, a temporally inconsistent perturbation, which sits at the left end of the consistency axis, is almost completely erased by every attack, as its retention drops to nearly zero. The reason is that per-frame independent perturbations are temporally uncorrelated. Frame-averaging operations act as low-pass filters along the temporal dimension, attenuating the adversarial perturbation in pixel space. This suggests that $\delta_y$ is encouraged to be temporally consistent to be robust against such attacks. More details on the definition of temporal attacks and experiments are provided in Appendix~\ref{app:rethinking-2}.

\section{\texttt{TC-UAP}: Temporally Consistent Universal Adversarial Perturbations}
\label{sec:method}

In this section, we propose \texttt{TC-UAP}, an identity-level Temporally Consistent Universal Adversarial Perturbation for video protection against both tuning-based and reference-based customization.

\subsection{Identity-Level Multi-Frame UAP}
\label{sec:uap}

\textbf{Identity-level multi-frame formulation.}
To enable practical and efficient protection, we seek to generate adversarial perturbations that can accommodate videos of varying lengths and generalize to unseen videos of the same identity without requiring per-video optimization. Accordingly, we model the protection signal as an identity-level universal adversarial perturbation (UAP). 
For identity $y$, given a training video set $\mathcal{V}_y^{\text{Tr}} =\{V_i\}_{i=1}^{M_{\text{Tr}}} \subseteq \mathcal{V}_y$, we learn a UAP $\delta_y$ shared across videos of that identity. To provide sufficient temporal expressiveness beyond conventional one-frame UAPs\footnote{Empirically, we find that a one-frame UAP, when repeated across video frames, fails to provide sufficiently strong video-level protection under the video VAE latent-disruption objective. Detailed results are provided in \cref{fig:uap_cap}.}\cite{moosavi2017universal}, 
we instantiate $\delta_y$ as a multi-frame UAP comprising $N$ temporally indexed perturbation frames:
\begin{equation}
    \delta_y \in \mathbb{R}^{N \times C \times H \times W},
    \qquad
    \|\delta_y\|_\infty \le \eta,
\end{equation}
where $\eta$ is the perturbation budget and we use $\delta_y^n$ to denote its $n$-th perturbation frame.
To protect a video $V_i$ of length $T_i$, we repeat $\delta_y$ along the temporal dimension until it covers all $T_i$ frames:
\begin{equation}
    \delta_{T_i}
    =
    \mathrm{Repeat}_{T_i}(\delta_y)
    =
    \underbrace{
    \big\{
    \delta_y^{1}, \dots, \delta_y^{N},
    \delta_y^{1}, \dots, \delta_y^{N},
    \delta_y^{1}, \dots
    \big\}
    }_{\text{repeated and cropped to } T_i \text{ frames}},
    \label{eq:repeat}
\end{equation}
and obtain the protected video as $V_i' = V_i + \delta_{T_i}$. This construction allows the fixed-length multi-frame UAP to be applied to videos of varying lengths.

\textbf{Video-level latent disruption objective.}
To provide unified protection against both customization pipelines, we target video VAE latent representations, which constitute a common intermediate representation in reference- and tuning-based customization. 
To mitigate the weakening effect of \textit{temporal compression} in 3D video VAEs, we optimize the perturbation jointly at the video level rather than frame by frame.
Therefore, an ideal full-video latent-disruption objective can be formulated as:
\begin{equation}
    \mathcal{L}_{\text{lat}}^{\text{full}}(\delta_y)
    =
    -\,\mathbb{E}_{V_i \sim \mathcal{V}_y^{\text{Tr}}}
    \left\|
    \mathcal{E}\big(V_i + \mathrm{Repeat}_{T_i}(\delta_y)\big)
    -
    \mathcal{E}(V_i)
    \right\|_2^2,
    \label{eq:lat-full}
\end{equation}
where $\mathcal{E}$ denotes the video VAE encoder. Minimizing this objective encourages the multi-frame UAP to induce a large distance between the latent
representations of the clean and protected videos.

However, directly optimizing Eq.~\eqref{eq:lat-full} over full-length videos incurs substantial \textit{memory overhead} and may cause the learned perturbation to \textit{overfit to the temporal sequences} of the training videos, including their full durations, fixed sequence boundaries, and fixed alignment between the causal sequence boundary and the starting phase of the repeated multi-frame UAP. Temporal edits such as starting-point shifts, cropping, and length variations alter these properties and may therefore weaken the protection.

\textbf{Sliding-window proxy.}
We therefore use sliding windows as a proxy for the full-video objective. 
As shown in \cref{fig:obs1b}(b) left column, under causal temporal modeling, each temporal latent position primarily depends on a local span of the current and preceding input frames. 
Accordingly, different latent positions within a window involve different effective history lengths, which gradually increase until the temporal receptive field is largely saturated.
Motivated by this observation, we choose the window length $L$ to cover at least the dominant portion of the temporal receptive field of the 3D causal video VAE. Although shorter than the full video, each $L$-length window captures sufficient temporal dependencies to account for the effects of temporal compression.

Formally, for a training video $V_i \in \mathcal{V}_y^{\text{Tr}}$ of length $T_i$ and its protected version $V_i' = V_i + \mathrm{Repeat}_{T_i}(\delta_y)$, we sample a starting temporal position $k \in \{0,\ldots,T_i-L\}$ and extract:
\begin{equation}
    X = \{v_i^j\}_{j=k}^{k+L-1},
    \qquad
    X' = \{v_i'^{j}\}_{j=k}^{k+L-1}.
    \label{eq:window}
\end{equation}
The practical latent-disruption objective is then defined as:
\begin{equation}
    \mathcal{L}_{\text{lat}}(\delta_y)
    =
    -\,\mathbb{E}_{V_i \sim \mathcal{V}_y^{\text{Tr}},\,k}
    \left\|
    \mathcal{E}(X') - \mathcal{E}(X)
    \right\|_2^2.
    \label{eq:lat}
\end{equation}
Intuitively, this sliding-window objective serves as an effective proxy for the full-video objective. For a video longer than the optimization window, each latent position still depends primarily on a local temporal context whose dominant receptive field can be covered by the $L$-length window. When the multi-frame UAP is temporally repeated across such a video, latent features at later temporal positions remain influenced by the adversarial perturbations within their corresponding receptive fields, similarly to how latent features are perturbed during window-based optimization. Consequently, perturbation patterns learned within fixed-length windows can be extended to longer videos through temporal repetition while remaining effective under longer-sequence encoding.

Taken together, window-based optimization not only reduces the substantial memory overhead of full-video optimization but also enables the multi-frame UAP to maintain its protection effectiveness under temporal edits.

\begin{algorithm}[t]
\caption{Training the identity-level multi-frame UAP $\delta_y$}
\label{alg:uap_train}
\begin{algorithmic}[1]
\Require training videos for an identity $y$: $\mathcal{V}_y^{\text{Tr}} = \{V_i\}_{i=1}^{M_{\text{Tr}}}$, multi-frame UAP length $N$, window length $L$, budget $\eta$, surrogate temporal attack $\mathcal{T}$, surrogate-loss weight $\lambda$, learning rate $\gamma$, epochs $E$, and $R$ from Eq.~\eqref{eq:residual}.
\State Initialize $\theta_y \sim \mathrm{Uniform}(-0.01, 0.01)$
\For{$e = 1$ \textbf{to} $E$}
    \For{each $V_i \in \mathcal{V}_y^{\text{Tr}}$}
        \State $\delta_y \gets R(\theta_y)$ \Comment{Eq.~\eqref{eq:residual}}
        \State $\delta_{T_i} \gets \mathrm{Repeat}_{T_i}(\delta_y)$ \Comment{Eq.~\eqref{eq:repeat}}
        \State $V_i' \gets V_i + \delta_{T_i}$
        \For{each window position $k$ sampled from $V_i$}
        \State $X \gets V_i[k : k + L-1],\quad X' \gets V_i'[k : k + L-1]$ \Comment{clean / protected window}
        \State $z \gets \mathcal{E}(X),\quad z' \gets \mathcal{E}(X')$
        \State $\mathcal{L}_{\text{lat}} \gets -\,\|z' - z\|_2^2$ 
        \State $\tilde{z} \gets \mathcal{E}(\mathcal{T}(X)),\quad \tilde{z}' \gets \mathcal{E}(\mathcal{T}(X'))$ 
        \State $\mathcal{L}_{\text{temporal}} \gets -\,\|\tilde{z}' - \tilde{z}\|_2^2$
        \State $\mathcal{L}_{\text{id}} \gets \mathcal{L}_{\text{lat}} + \lambda \cdot \mathcal{L}_{\text{temporal}}$ \Comment{overall objective}
        \State $\theta_y \gets \mathrm{Adam}\!\left(\theta_y,\, \nabla_{\theta_y}\mathcal{L}_{\text{id}},\, \gamma\right)$ \Comment{update $\theta_y$}
        \State $\delta_y \gets R(\theta_y)$ 
        \State $\delta_y \gets \mathrm{clip}\!\left(\delta_y,\, [-\eta, \eta]\right);\ \ \theta_y \gets R^{-1}(\delta_y)$ \Comment{$\ell_\infty$ projection in $\delta_y$-space}
        \EndFor
    \EndFor
\EndFor
\State \Return $\delta_y \gets R(\theta_y)$
\end{algorithmic}
\end{algorithm}

\subsection{Temporal Consistency Design} \label{sec:temporal}

Although the multi-frame formulation provides greater temporal expressiveness, a multi-frame UAP is not necessarily temporally consistent when its frames are independently parameterized, leaving the resulting perturbation vulnerable to temporal attacks. 
To address this issue, we improve temporal robustness from two complementary aspects: 
an extrinsic surrogate temporal-attack loss and intrinsic temporal consistency modeling.

\textbf{Extrinsic surrogate temporal-attack loss.}
We first introduce a surrogate temporal-attack loss that directly optimizes the multi-frame UAP 
to remain effective after temporal attack:
\begin{equation}
    \mathcal{L}_{\text{temporal}}(\delta_y)
    =
    -\,\mathbb{E}_{X,\,X'}
    \big\|\,\mathcal{E}(\mathcal{T}(X')) - \mathcal{E}(\mathcal{T}(X))\,\big\|_2^2,
    \label{eq:tmp}
\end{equation}
where $\mathcal{T}$ is a surrogate temporal attack, such as frame averaging, which averages each pair of adjacent frames with weights $0.5/0.5$.
Specifically, this objective encourages the multi-frame UAP to still preserve a large VAE 
latent distance after the surrogate temporal attack, thus explicitly enhancing its robustness. 
However, such attack-aware optimization may become tied to the specific surrogate attack and fail to generalize to unseen temporal attacks.

\textbf{Intrinsic temporal consistency modeling.}
As illustrated in \cref{fig:obs3}, perturbations with stronger temporal consistency retain their attack strength more effectively across different temporal attacks. Motivated by this observation, we introduce an \textit{attack-agnostic temporal-consistency prior} into the multi-frame UAP parameterization to improve generalization to unseen temporal attacks. 
The parameterization is designed to balance temporal consistency with perturbation expressiveness, as overly strong temporal constraints may limit the expressiveness required for video protection, whereas weak temporal correlation may leave different frames nearly independent, making stable temporal consistency difficult to achieve.
Based on this trade-off, we use a reparameterization based on the average of previous frames. 
Specifically, we introduce a parameter 
$\theta_y \in \mathbb{R}^{N \times C \times H \times W}$ and define the reparameterization $\delta_y = R(\theta_y)$ as:
\begin{equation}
    \delta_y^{1} \;=\; \theta_y^{1}, \qquad
    \delta_y^{n} \;=\; \frac{1}{n-1}\sum_{i=1}^{n-1}\delta_y^{i} \;+\; \theta_y^{n},
    \quad \text{for } n \ge 2,
    \label{eq:residual}
\end{equation}
with inverse $\theta_y = R^{-1}(\delta_y)$, where $\delta_y^n$ and $\theta_y^n$ denote the $n$-th frame of $\delta_y$ and $\theta_y$, respectively.

In this formulation, each UAP frame consists of two terms: the average of all previous UAP frames and a learnable frame-specific term. The first term makes $\delta_y^n$ depend on perturbations at earlier temporal indices, thereby introducing temporal correlation across perturbation frames. The second term $\theta_y^n$ retains a learnable component specific to the $n$-th frame, thereby preserving the expressiveness of the multi-frame UAP. This attack-agnostic temporal structure complements the surrogate temporal-attack loss and improves robustness against unseen temporal attacks.

\subsection{Optimization for \texttt{TC-UAP}}\label{app:algo}

\textbf{Overall objective.}
Combining the untargeted latent distance loss in Eq.~\eqref{eq:lat} with the surrogate temporal attack loss in Eq.~\eqref{eq:tmp}, the total training objective is:
\begin{equation}
    \mathcal{L}_{\text{id}}(\delta_y) \;=\; \mathcal{L}_{\text{lat}}(\delta_y) \;+\; \lambda\,\mathcal{L}_{\text{temporal}}(\delta_y),\quad\delta_y = R(\theta_y),
    \label{eq:total-loss}
\end{equation}
where $\lambda$ controls the contribution of the surrogate temporal-attack loss. Minimizing $\mathcal{L}_{\text{id}}$ encourages the multi-frame UAP to induce a large VAE latent distance both before and after the surrogate temporal attack, thereby jointly optimizing protection effectiveness and temporal robustness.

\textbf{Sliding-window-based Projected Gradient Descent.}
For each identity $y$, the UAP is optimized over a training video set $\mathcal{V}_y^{\text{Tr}} = \{V_i\}_{i=1}^{M_{\text{Tr}}}$.
At each training step, we sample a video $V_i \in \mathcal{V}_y^{\text{Tr}}$ with length $T_i$, and construct the corresponding protected video $V_i' = V_i + \mathrm{Repeat}_{T_i}(\delta_y)$ according to Eq.~\eqref{eq:repeat}.
We then uniformly sample a starting frame index $k \in \{0,\dots,T_i-L\}$, and extract the clean and protected windows from the same temporal position via Eq.~\eqref{eq:window}, i.e., $X = \{v_i^j\}_{j=k}^{k+L-1}$ and $X' = \{v_i'^{j}\}_{j=k}^{k+L-1}$.
The total loss is then computed according to Eq.~\eqref{eq:total-loss}.
We optimize $\theta_y$ by back-propagating gradients through $\delta_y = R(\theta_y)$.
After each update, the UAP $\delta_y = R(\theta_y)$ is projected in $\delta_y$-space to satisfy the $\ell_\infty$ budget:
\begin{equation}
    \delta_y \leftarrow
    \Pi_{\|\delta_y\|_\infty \le \eta}\big(R(\theta_y)\big),
    \qquad
    \theta_y \leftarrow R^{-1}(\delta_y),
    \label{eq:projection}
\end{equation}
where $\Pi_{\|\delta_y\|_\infty \le \eta}$ denotes the element-wise projection onto the $\ell_\infty$ ball of radius $\eta$, implemented by clamping each perturbation value to $[-\eta,\eta]$.
\textit{The complete optimization procedure is summarized in Algorithm~\ref{alg:uap_train}.}

\textbf{Applying the learned UAP to unseen videos.}
After optimization, given a new video $V$ with length $T$, the protected video $V'$ can be obtained by repeating $\delta_y$ along the temporal dimension until it covers all frames, and cropping it to length $T$:
\begin{equation}
    V' = V + \mathrm{Repeat}_T(\delta_y),
\end{equation}
where additional optimization for each new video is not required. Therefore, the identity-level multi-frame UAP provides generalizable and efficient protection, as it can be reused across unseen videos of the same identity.

\begin{figure*}[b]
    \centering
    \includegraphics[width=\textwidth]{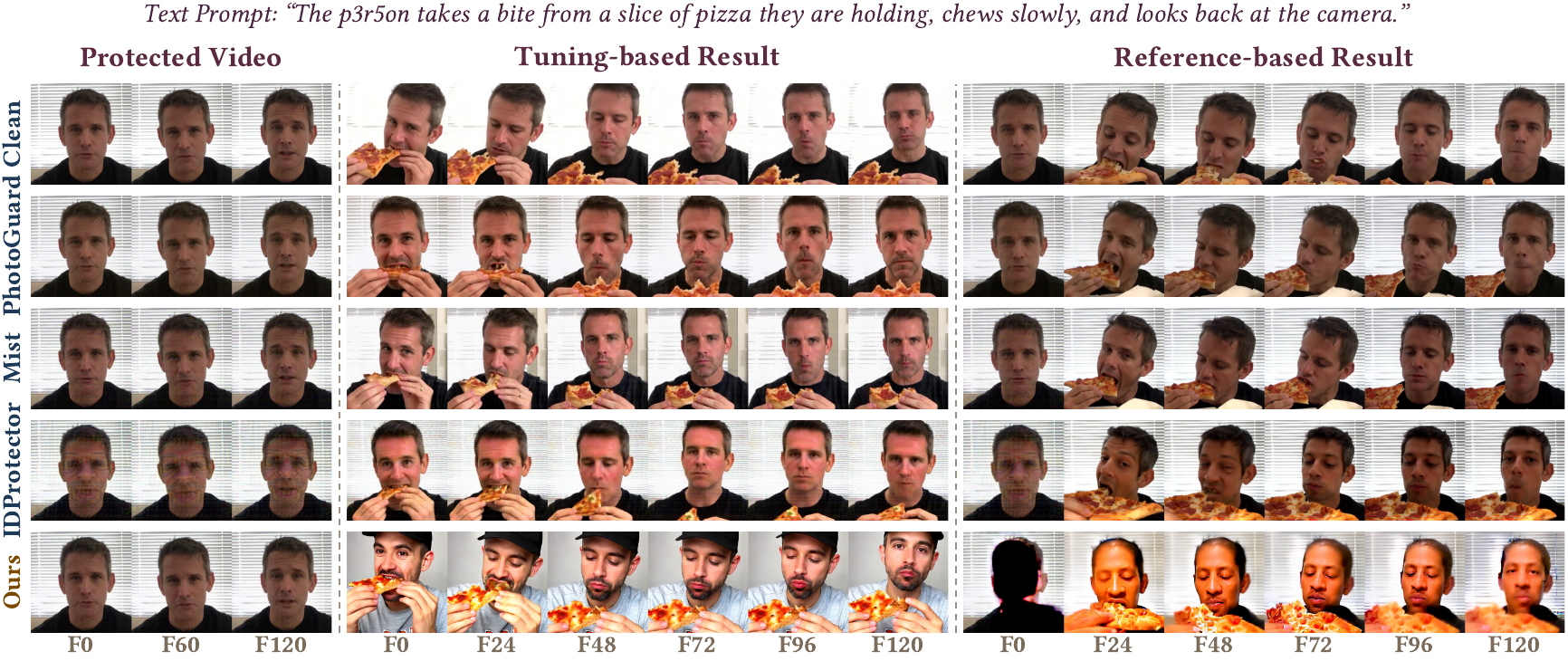}
    \caption{Qualitative comparison of reference- and tuning-based customization results.}
    \label{fig:main}                                                        
\end{figure*}
\section{Experiments} \label{sec:experiments}

\subsection{Experimental Setup}

\textbf{Datasets. } 
To the best of our knowledge, there is no existing benchmark for protecting videos against diffusion-based customization. Therefore, we adapt three human-centered video datasets and evaluate on them: HDTF~\cite{zhang2021flow}, CelebV-HQ~\cite{zhu2022celebvhq} and TalkVid~\cite{chen2025talkvid}. For each dataset, we select 20 identities and split videos of each identity into 15 training clips for UAP optimization and 15 test clips for evaluating protection effectiveness. More dataset details and identity selection criteria are provided in Appendix~\ref{app:dataset-details}.

\textbf{Evaluation metrics. }
Our method is evaluated from two perspectives: protection effectiveness and invisibility.
For protection effectiveness, we report VAE reconstruction metrics including PSNR, SSIM~\cite{wang2004image} and LPIPS~\cite{zhang2018perceptual}; and identity preservation metrics, including Face Detection Failure Rate (FDFR)~\cite{Deng_2020_CVPR} and identity similarity measured by ArcFace~\cite{Deng_2019_CVPR} (ISM1) and CurricularFace~\cite{huang2020curricularface} (ISM2).
For invisibility, we report Video Multi-Method Assessment Fusion (VMAF)~\cite{netflix2video}. Higher VMAF indicates better perceptual quality and less visible perturbation. Detailed definitions are provided in Appendix~\ref{app:eval-details}.

\textbf{Implementation. }
Following prior image protection works against diffusion-based customization~\cite{pmlr-v202-salman23a, wang2024simac, van2023anti, liang2023mist, liang2023adversarial}, we adopt a white-box setting in which the protector has access to the VAE encoder.
For each identity, we set the UAP length to 9 frames and optimize it on 15 training clips with perturbation budget $\eta=0.1$. Following Eq.~\eqref{eq:repeat}, the learned UAP is temporally repeated and added to each test clip to obtain the protected test set, which consists of 15 test clips.
For tuning-based evaluation, we fine-tune LoRA adapters for LTX-2.3~\cite{hacohen2026ltx2efficientjointaudiovisual} and Wan2.2-5B~\cite{wan2025wanopenadvancedlargescale} on the protected test set.
For reference-based evaluation, the official LTX-2.3 and Wan2.2-5B TI2V pipelines are used, each conditioned on an extracted frame of a protected video. The surrogate temporal attack $\mathcal{T}$ is implemented as frame averaging, in which adjacent frames are averaged with weights $0.5/0.5$.
More details are deferred to Appendix~\ref{app:impl}.

\textbf{Baselines. } We consider three baseline image protection methods: PhotoGuard~\cite{pmlr-v202-salman23a}, Mist~\cite{liang2023mist}, IDProtector~\cite{song2025idprotector}.
We apply each baseline to every frame of the test videos and adopt the same perturbation budget $\eta=0.1$.

\begin{table*}[t]
    \centering
    \caption{Results of protection quality evaluated under tuning-based and reference-based settings.
    \textbf{Bold} indicates the best result and \underline{underline} the second-best.
    On the \emph{Ours} row, \textcolor{green!50!black}{green} percentages report the relative improvement of our method against the strongest baseline.}
    \label{tab:main}

    \footnotesize
    \setlength{\tabcolsep}{3.2pt}
    \renewcommand{\arraystretch}{1.12}

    \resizebox{\textwidth}{!}{%
    \begin{tabular}{llccc ccc ccc ccc}
    \toprule
    \multirow{3}{*}{\textbf{Dataset}} &
    \multirow{3}{*}{\textbf{Methods}} &
    \multicolumn{6}{c}{\textbf{Tuning-based}} &
    \multicolumn{6}{c}{\textbf{Reference-based}} \\
    \cmidrule(lr){3-8}
    \cmidrule(lr){9-14}
    & &
    \multicolumn{3}{c}{\textbf{VAE Reconstruction}} &
    \multicolumn{3}{c}{\textbf{Identity Preservation}} &
    \multicolumn{3}{c}{\textbf{VAE Reconstruction}} &
    \multicolumn{3}{c}{\textbf{Identity Preservation}} \\
    \cmidrule(lr){3-5}
    \cmidrule(lr){6-8}
    \cmidrule(lr){9-11}
    \cmidrule(lr){12-14}
    & &
    PSNR\,$\!\downarrow$ &
    SSIM\,$\!\downarrow$ &
    LPIPS\,$\!\uparrow$ &
    FDFR\,$\!\uparrow$ &
    ISM1\,$\!\downarrow$ &
    ISM2\,$\!\downarrow$ &
    PSNR\,$\!\downarrow$ &
    SSIM\,$\!\downarrow$ &
    LPIPS\,$\!\uparrow$ &
    FDFR\,$\!\uparrow$ &
    ISM1\,$\!\downarrow$ &
    ISM2\,$\!\downarrow$ \\
    \midrule

    \multirow{6}{*}{\textbf{HDTF}}
    & Clean
    & 39.64 & 0.973 & 0.016
    & 0.000 & 0.598 & 0.609
    & 41.69 & 0.979 & 0.011
    & 0.017 & 0.607 & 0.624 \\

    & PhotoGuard
    & 29.44 & \underline{0.618} & \underline{0.590}
    & 0.000 & 0.572 & 0.585
    & 30.64 & \underline{0.688} & \underline{0.451}
    & \underline{0.020} & 0.556 & 0.573 \\

    & Mist
    & 31.54 & 0.770 & 0.347
    & 0.000 & 0.544 & 0.557
    & 31.25 & 0.743 & 0.374
    & 0.003 & 0.527 & 0.544 \\

    & IDProtector
    & \underline{29.15} & 0.757 & 0.397
    & 0.000 & \underline{0.430} & \underline{0.459}
    & \underline{29.33} & 0.763 & 0.401
    & 0.003 & \underline{0.179} & \underline{0.191} \\
    \addlinespace[1.5pt]

    \rowcolor{orange!8}
    \cellcolor{white} & 
    & \textbf{13.86} & \textbf{0.368} & \textbf{0.657}
    & \textbf{0.027} & \textbf{0.077} & \textbf{0.080}
    & \textbf{10.08} & \textbf{0.153} & \textbf{0.736}
    & \textbf{0.321} & \textbf{0.129} & \textbf{0.141} \\[-2pt]
    \rowcolor{orange!8}
    \cellcolor{white} & \multirow{-1.7}{*}{Ours}
    & {\tiny\textcolor{green!50!black}{$\uparrow$52.45\%}}
    & {\tiny\textcolor{green!50!black}{$\uparrow$40.45\%}}
    & {\tiny\textcolor{green!50!black}{$\uparrow$11.36\%}}
    & {\tiny\textcolor{gray!70}{--}}
    & {\tiny\textcolor{green!50!black}{$\uparrow$82.09\%}}
    & {\tiny\textcolor{green!50!black}{$\uparrow$82.57\%}}
    & {\tiny\textcolor{green!50!black}{$\uparrow$65.63\%}}
    & {\tiny\textcolor{green!50!black}{$\uparrow$77.76\%}}
    & {\tiny\textcolor{green!50!black}{$\uparrow$63.19\%}}
    & {\tiny\textcolor{green!50!black}{$\uparrow$1505.00\%}}
    & {\tiny\textcolor{green!50!black}{$\uparrow$27.93\%}}
    & {\tiny\textcolor{green!50!black}{$\uparrow$26.18\%}} \\

    \midrule

    \multirow{6}{*}{\textbf{CelebV-HQ}}
    & Clean
    & 39.47 & 0.972 & 0.019
    & 0.000 & 0.630 & 0.650
    & 42.27 & 0.981 & 0.011
    & 0.013 & 0.611 & 0.637 \\

    & PhotoGuard
    & 29.26 & \underline{0.617} & \underline{0.600}
    & 0.000 & 0.599 & 0.617
    & 30.59 & \underline{0.695} & \underline{0.464}
    & \underline{0.017} & 0.543 & 0.571 \\

    & Mist
    & 31.35 & 0.767 & 0.378
    & 0.000 & 0.554 & 0.579
    & 31.23 & 0.743 & 0.398
    & 0.009 & 0.501 & 0.525 \\

    & IDProtector
    & \underline{29.16} & 0.757 & 0.413
    & 0.000 & \underline{0.446} & \underline{0.475}
    & \underline{29.43} & 0.764 & 0.418
    & 0.013 & \underline{0.232} & \underline{0.250} \\
    \addlinespace[1.5pt]

    \rowcolor{orange!8}
    \cellcolor{white} & 
    & \textbf{12.91} & \textbf{0.374} & \textbf{0.645}
    & \textbf{0.006} & \textbf{0.113} & \textbf{0.125}
    & \textbf{8.84} & \textbf{0.171} & \textbf{0.708}
    & \textbf{0.225} & \textbf{0.139} & \textbf{0.140} \\[-2pt]
    \rowcolor{orange!8}
    \cellcolor{white} & \multirow{-1.7}{*}{Ours}
    & {\tiny\textcolor{green!50!black}{$\uparrow$55.73\%}}
    & {\tiny\textcolor{green!50!black}{$\uparrow$39.38\%}}
    & {\tiny\textcolor{green!50!black}{$\uparrow$7.50\%}}
    & {\tiny\textcolor{gray!70}{--}}
    & {\tiny\textcolor{green!50!black}{$\uparrow$74.66\%}}
    & {\tiny\textcolor{green!50!black}{$\uparrow$73.68\%}}
    & {\tiny\textcolor{green!50!black}{$\uparrow$69.96\%}}
    & {\tiny\textcolor{green!50!black}{$\uparrow$75.40\%}}
    & {\tiny\textcolor{green!50!black}{$\uparrow$52.59\%}}
    & {\tiny\textcolor{green!50!black}{$\uparrow$1223.53\%}}
    & {\tiny\textcolor{green!50!black}{$\uparrow$40.09\%}}
    & {\tiny\textcolor{green!50!black}{$\uparrow$44.00\%}} \\

    \midrule

    \multirow{6}{*}{\textbf{TalkVid}}
    & Clean
    & 36.94 & 0.960 & 0.018
    & 0.000 & 0.623 & 0.636
    & 39.19 & 0.971 & 0.013
    & 0.000 & 0.628 & 0.650 \\

    & PhotoGuard
    & \underline{28.51} & \underline{0.602} & \underline{0.567}
    & 0.000 & 0.585 & 0.598
    & 29.97 & \underline{0.691} & \underline{0.438}
    & \underline{0.001} & 0.572 & 0.595 \\

    & Mist
    & 30.44 & 0.761 & 0.349
    & 0.000 & 0.551 & 0.568
    & 30.24 & 0.734 & 0.380
    & \underline{0.001} & 0.527 & 0.545 \\

    & IDProtector
    & 28.69 & 0.751 & 0.395
    & 0.000 & \underline{0.431} & \underline{0.463}
    & \underline{29.10} & 0.765 & 0.403
    & \underline{0.001} & \underline{0.310} & \underline{0.327} \\
    \addlinespace[1.5pt]

    \rowcolor{orange!8}
    \cellcolor{white} & 
    & \textbf{12.17} & \textbf{0.381} & \textbf{0.637}
    & \textbf{0.013} & \textbf{0.160} & \textbf{0.164}
    & \textbf{8.00} & \textbf{0.187} & \textbf{0.709}
    & \textbf{0.146} & \textbf{0.275} & \textbf{0.286} \\[-2pt]
    \rowcolor{orange!8}
    \cellcolor{white} & \multirow{-1.7}{*}{Ours}
    & {\tiny\textcolor{green!50!black}{$\uparrow$57.31\%}}
    & {\tiny\textcolor{green!50!black}{$\uparrow$36.71\%}}
    & {\tiny\textcolor{green!50!black}{$\uparrow$12.35\%}}
    & {\tiny\textcolor{gray!70}{--}}
    & {\tiny\textcolor{green!50!black}{$\uparrow$62.88\%}}
    & {\tiny\textcolor{green!50!black}{$\uparrow$64.58\%}}
    & {\tiny\textcolor{green!50!black}{$\uparrow$72.51\%}}
    & {\tiny\textcolor{green!50!black}{$\uparrow$72.94\%}}
    & {\tiny\textcolor{green!50!black}{$\uparrow$61.87\%}}
    & {\tiny\textcolor{gray!70}{--}}
    & {\tiny\textcolor{green!50!black}{$\uparrow$11.29\%}}
    & {\tiny\textcolor{green!50!black}{$\uparrow$12.54\%}} \\

    \bottomrule
    \end{tabular}%
    }
\end{table*}
\subsection{Main Results}
\begin{wraptable}{r}{0.5\columnwidth}
    \centering
    \vspace{-3\baselineskip}
    \caption{Invisibility of the perturbation (VMAF\,$\!\uparrow$).}
    \label{tab:invisibility}
    \footnotesize
    \setlength{\tabcolsep}{6pt}
    \renewcommand{\arraystretch}{1.12}
    \resizebox{0.5\columnwidth}{!}{%
    \begin{tabular}{lccc}
    \toprule
    \textbf{Method} & \textbf{HDTF} & \textbf{CelebV-HQ} & \textbf{TalkVid} \\
    \midrule
    \textit{Clean}                      & 99.17 & 100.64 & 100.14 \\
    \textit{Random ($\varepsilon=0.1$)} & 92.65 & 93.87  & 94.01  \\
    \midrule
    PhotoGuard   & \underline{80.86} & \underline{82.76} & \underline{81.93} \\
    Mist         & 74.63 & 76.76 & 75.49 \\
    IDProtector  & 52.53 & 54.94 & 56.50 \\
    \addlinespace[1.5pt]
    \rowcolor{orange!8}
    Ours
    & \textbf{82.76}\,{\tiny\textcolor{green!50!black}{(+1.90)}}
    & \textbf{84.60}\,{\tiny\textcolor{green!50!black}{(+1.84)}}
    & \textbf{84.81}\,{\tiny\textcolor{green!50!black}{(+2.88)}} \\
    \bottomrule
    \end{tabular}%
    }
\end{wraptable}

We compare our method against the baselines across three datasets on LTX-2.3~\cite{hacohen2026ltx2efficientjointaudiovisual} under both the reference- and tuning-based customization pipelines. Quantitative results are reported in \cref{tab:main}. Across all datasets, our method achieves the lowest ISM and the highest FDFR, indicating that the generated identity differs most from the target identity and is even undetectable as a face in many frames. Besides, as shown in \cref{tab:invisibility}, our method achieves the highest VMAF, indicating that our protection is the most imperceptible. 

We also analyze VAE reconstruction to verify whether the input latent is effectively disrupted, as shown in \cref{tab:main}. Since the two customization pipelines take different forms of input, we evaluate reconstruction in two settings. For tuning-based customization, the input is a set of videos, so we measure video-level VAE reconstruction. For reference-based customization, the input is a reference image, so we measure image-level VAE reconstruction. Compared with the strongest baseline, our method reduces reconstruction PSNR by 15.96 dB in the video-level setting and 20.31 dB in the image-level setting. The corresponding SSIM drops are 0.24 and 0.52, respectively. These results show that our method strongly disrupts the VAE latent, which is consistent with its strong ability to reduce identity preservation in customization.

We also present qualitative results in \cref{fig:main}. Under our method, the generated faces no longer look like the target person, while most baselines still produce faces that clearly look like the target. Among them, IDProtector shows some degree of protection in reference-based customization results. However, IDProtector adds visible patterns at fixed facial regions, which are clearly visible. More qualitative comparisons are shown in Appendix~\ref{app:qualitative}.

\begin{figure}[t]
    \centering
    \includegraphics[width=0.85\textwidth]{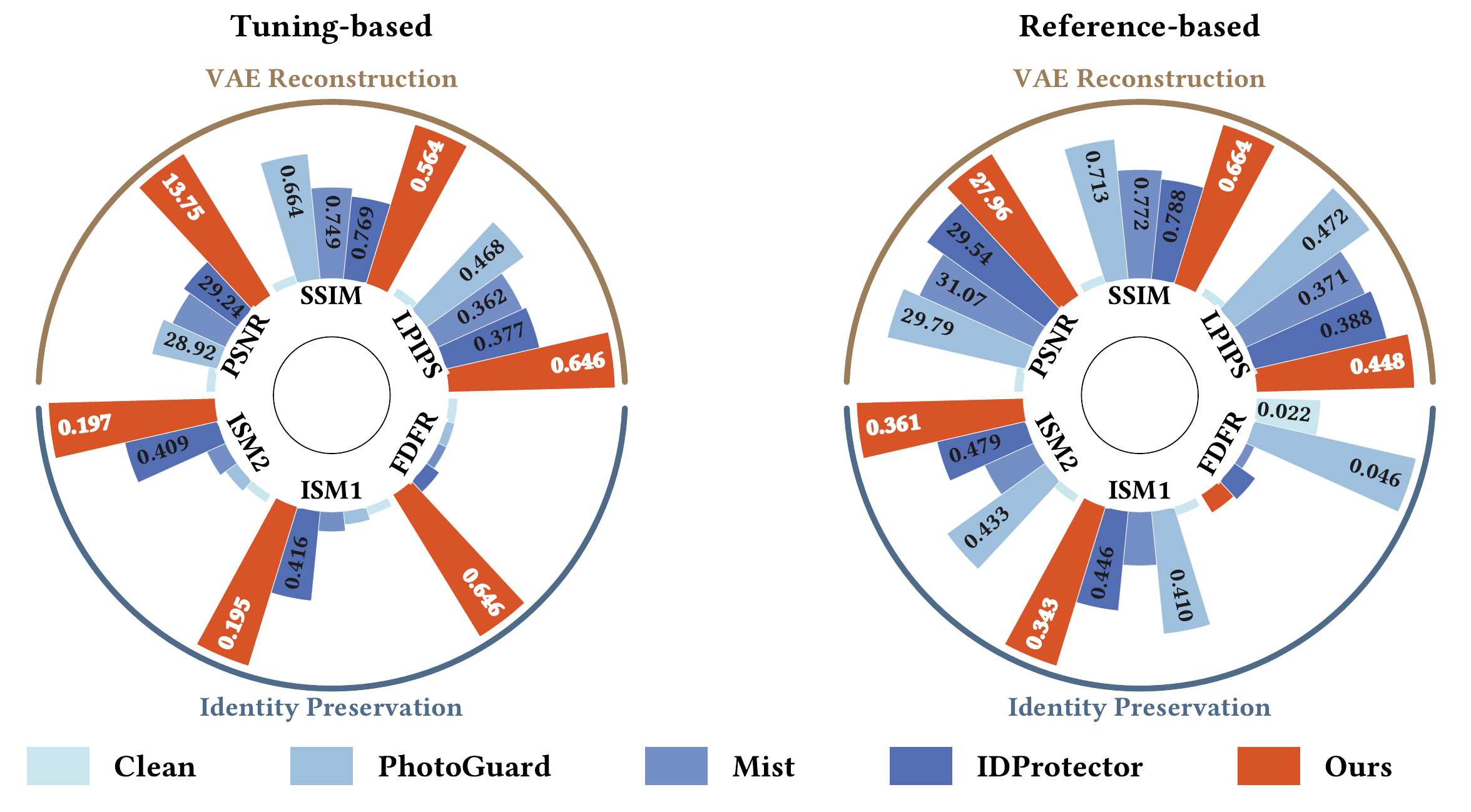}
    \caption{Protection quality on the TalkVid dataset using the Wan2.2 video diffusion model~\cite{wan2025wanopenadvancedlargescale}.}
    \label{fig:wan22}
\end{figure}

In addition, we evaluate our method on the Wan2.2 video diffusion model~\cite{wan2025wanopenadvancedlargescale} using the TalkVid dataset. Results are shown in \cref{fig:wan22}. Under both customization pipelines, our method again achieves the lowest VAE reconstruction quality and the lowest ISM scores overall. Under the reference-based pipeline, PhotoGuard achieves slightly higher LPIPS and FDFR than our method. However, on the identity-preservation metrics, our method still outperforms.

\subsection{Robustness evaluation against temporal attacks}
\begin{table*}[htbp]
    \centering
    \caption{Robustness results under different unseen temporal attacks on the TalkVid dataset.}
    \label{tab:robustness}
    \footnotesize
    \setlength{\tabcolsep}{3.2pt}
    \renewcommand{\arraystretch}{1.2}
    \resizebox{\textwidth}{!}{%
    \begin{tabular}{llccc ccc ccc ccc}
    \toprule
    \multirow{3}{*}{\textbf{Operation}} &
    \multirow{3}{*}{\textbf{Methods}} &
    \multicolumn{6}{c}{\textbf{Tuning-based}} &
    \multicolumn{6}{c}{\textbf{Reference-based}} \\
    \cmidrule(lr){3-8} \cmidrule(lr){9-14}
    & & \multicolumn{3}{c}{\textbf{VAE Reconstruction}} & \multicolumn{3}{c}{\textbf{Identity Preservation}} & \multicolumn{3}{c}{\textbf{VAE Reconstruction}} & \multicolumn{3}{c}{\textbf{Identity Preservation}} \\
    \cmidrule(lr){3-5} \cmidrule(lr){6-8} \cmidrule(lr){9-11} \cmidrule(lr){12-14}
    & & PSNR$\downarrow$ & SSIM$\downarrow$ & LPIPS$\uparrow$ & FDFR$\uparrow$ & ISM1$\downarrow$ & ISM2$\downarrow$ & PSNR$\downarrow$ & SSIM$\downarrow$ & LPIPS$\uparrow$ & FDFR$\uparrow$ & ISM1$\downarrow$ & ISM2$\downarrow$ \\
    \midrule

    & PGD
    & \textbf{6.19} & \textbf{0.096} & \textbf{0.857}
    & \textbf{0.169} & \textbf{0.001} & \textbf{0.006}
    & 24.43 & 0.638 & 0.446
    & 0.001 & 0.534 & 0.551 \\
    \rowcolor{orange!8} 
    \cellcolor{white}\multirow{-2}{*}{\textbf{Original}} & Ours
    & 12.17 & 0.381 & 0.637
    & 0.013 & 0.160 & 0.164
    & \textbf{8.00} & \textbf{0.187} & \textbf{0.709}
    & \textbf{0.146} & \textbf{0.275} & \textbf{0.286} \\
    \midrule

    & PGD
    & 26.21 & 0.741 & 0.351
    & 0.000 & 0.463 & 0.481
    & 34.43 & 0.849 & 0.282
    & 0.004 & 0.550 & 0.576 \\
    \rowcolor{orange!8}
    \cellcolor{white}\multirow{-2}{*}{\textbf{Box Filter}} & Ours
    & \textbf{14.66} & \textbf{0.443} & \textbf{0.647}
    & \textbf{0.001} & \textbf{0.136} & \textbf{0.141}
    & \textbf{8.92} & \textbf{0.223} & \textbf{0.708}
    & \textbf{0.168} & \textbf{0.151} & \textbf{0.156} \\
    \midrule

    & PGD
    & 32.82 & 0.803 & 0.356
    & 0.000 & 0.469 & 0.491
    & 32.37 & 0.806 & 0.342
    & 0.002 & 0.540 & 0.561 \\
    \rowcolor{orange!8}
    \cellcolor{white}\multirow{-2}{*}{\textbf{Gaussian Filter}} & Ours
    & \textbf{14.49} & \textbf{0.432} & \textbf{0.681}
    & \textbf{0.013} & \textbf{0.111} & \textbf{0.117}
    & \textbf{8.91} & \textbf{0.217} & \textbf{0.715}
    & \textbf{0.165} & \textbf{0.164} & \textbf{0.171} \\
    \midrule

    & PGD
    & 34.66 & 0.861 & 0.280
    & 0.000 & 0.503 & 0.523
    & 36.64 & 0.905 & 0.216
    & 0.002 & 0.361 & 0.389 \\
    \rowcolor{orange!8}
    \cellcolor{white}\multirow{-2}{*}{\textbf{Low-pass Filter}} & Ours
    & \textbf{14.23} & \textbf{0.426} & \textbf{0.685}
    & \textbf{0.017} & \textbf{0.097} & \textbf{0.107}
    & \textbf{8.57} & \textbf{0.197} & \textbf{0.754}
    & \textbf{0.220} & \textbf{0.077} & \textbf{0.071} \\

    \bottomrule
    \end{tabular}%
    }
\end{table*}

We further evaluate the robustness of our method against temporal attacks. We test three unseen temporal attacks on the protected videos: box filter, Gaussian filter, and low-pass filter. The details of the three temporal attacks are shown in Appendix~\ref{app:eval-details}. 

We compare our method with a PGD baseline that directly optimizes a per-video perturbation for each video against the VAE encoder, without any temporal-consistency design, as in Eq.~\eqref{eq:per-video-baseline}. This gives strong protection on each individual video. Results are shown in \cref{tab:robustness}. On the original protected videos, the PGD baseline achieves the strongest protection against tuning-based customization. However, once any attack is applied, the PGD baseline loses most of its protection. PSNR rises by more than 20~dB after each attack, and ISM1 rises from $0.001$ to around $0.5$, meaning the protection is fully erased. In contrast, our method changes slightly after each filter. The tuning-based PSNR rises by less than 3~dB, and ISM1 stays around $0.10$. The reference-based metrics follow the same trend. These results demonstrate that our method is robust to unseen temporal attacks. This confirms that our temporal-consistency design provides robustness, rather than resistance to only the specific attack seen during training.

\subsection{Ablation Study}
\begin{table*}[htbp]
    \centering
    \caption{Ablation study on TalkVid dataset. 
    \textbf{ITM}: Intrinsic Temporal Modeling reparameterization (Eq.~\ref{eq:residual});
    \textbf{STA}: Surrogate Temporal-Attack loss (Eq.~\ref{eq:tmp}).
    \cmark/\xmark{} indicates whether the corresponding module is enabled.}
    \label{tab:ablation}
    \footnotesize
    \setlength{\tabcolsep}{3.2pt}
    \renewcommand{\arraystretch}{0.9}
    \resizebox{\textwidth}{!}{%
    \begin{tabular}{lcc ccc ccc ccc ccc}
    \toprule
    \multirow{3}{*}[-0.7\normalbaselineskip]{\textbf{Operation}} &
    \multirow{3}{*}[-0.7\normalbaselineskip]{\textbf{ITM}} &
    \multirow{3}{*}[-0.7\normalbaselineskip]{\textbf{STA}} &
    \multicolumn{6}{c}{\textbf{Tuning-based}} &
    \multicolumn{6}{c}{\textbf{Reference-based}} \\
    \cmidrule(lr){4-9}
    \cmidrule(lr){10-15}
    & & &
    \multicolumn{3}{c}{\textbf{VAE Reconstruction}} &
    \multicolumn{3}{c}{\textbf{Identity Preservation}} &
    \multicolumn{3}{c}{\textbf{VAE Reconstruction}} &
    \multicolumn{3}{c}{\textbf{Identity Preservation}} \\
    \cmidrule(lr){4-6}\cmidrule(lr){7-9}\cmidrule(lr){10-12}\cmidrule(lr){13-15}
    & & &
    PSNR\,$\!\downarrow$ & SSIM\,$\!\downarrow$ & LPIPS\,$\!\uparrow$ &
    FDFR\,$\!\uparrow$  & ISM1\,$\!\downarrow$ & ISM2\,$\!\downarrow$ &
    PSNR\,$\!\downarrow$ & SSIM\,$\!\downarrow$ & LPIPS\,$\!\uparrow$ &
    FDFR\,$\!\uparrow$  & ISM1\,$\!\downarrow$ & ISM2\,$\!\downarrow$ \\
    \midrule

    \multirow{3}{*}{\textbf{Original}}
    & \cmark & \xmark
    & 11.46 & 0.318 & 0.682
    & 0.003 & 0.061 & 0.071
    & 8.08  & 0.191 & 0.708
    & 0.084 & \textbf{0.232} & \textbf{0.240} \\
    & \xmark & \cmark
    & \textbf{6.47} & \textbf{0.082} & \textbf{0.826}
    & 0.000 & \textbf{0.010} & \textbf{0.017}
    & 14.08 & 0.358 & 0.609
    & 0.008 & 0.415 & 0.425 \\
    \rowcolor{orange!8}
    \cellcolor{white} & \cmark & \cmark
    & 12.17 & 0.381 & 0.637
    & \textbf{0.013} & 0.160 & 0.164
    & \textbf{8.00} & \textbf{0.187} & \textbf{0.709}
    & \textbf{0.146} & 0.275 & 0.286 \\
    \midrule

    \multirow{3}{*}{\textbf{Box Filter}}
    & \cmark & \xmark
    & 15.30 & 0.444 & 0.639
    & \textbf{0.013} & 0.173 & 0.182
    & 9.18 & 0.224 & 0.688
    & 0.111 & 0.162 & 0.173 \\
    & \xmark & \cmark
    & 18.50 & 0.501 & 0.574
    & 0.000 & 0.224 & 0.239
    & 14.93 & 0.415 & 0.607
    & 0.056 & 0.320 & 0.333 \\
    \rowcolor{orange!8}
    \cellcolor{white} & \cmark & \cmark
    & \textbf{14.66} & \textbf{0.443} & \textbf{0.647}
    & 0.001 & \textbf{0.136} & \textbf{0.141}
    & \textbf{8.92} & \textbf{0.223} & \textbf{0.708}
    & \textbf{0.168} & \textbf{0.151} & \textbf{0.156} \\
    \midrule

    \multirow{3}{*}{\textbf{Gaussian Filter}}
    & \cmark & \xmark
    & 15.36 & 0.437 & 0.656
    & 0.013 & 0.167 & 0.181
    & 9.17 & 0.221 & 0.687
    & 0.105 & \textbf{0.152} & \textbf{0.167} \\
    & \xmark & \cmark
    & 20.96 & 0.599 & 0.524
    & 0.000 & 0.313 & 0.327
    & 15.59 & 0.434 & 0.599
    & 0.050 & 0.358 & 0.372 \\
    \rowcolor{orange!8}
    \cellcolor{white} & \cmark & \cmark
    & \textbf{14.49} & \textbf{0.432} & \textbf{0.681}
    & \textbf{0.013} & \textbf{0.111} & \textbf{0.117}
    & \textbf{8.91} & \textbf{0.217} & \textbf{0.715}
    & \textbf{0.165} & 0.164 & 0.171 \\
    \midrule

    \multirow{3}{*}{\textbf{Low-pass Filter}}
    & \cmark & \xmark
    & 15.05 & 0.436 & 0.657
    & 0.006 & 0.165 & 0.177
    & 8.95 & 0.201 & 0.731
    & 0.201 & 0.093 & 0.099 \\
    & \xmark & \cmark
    & 21.35 & 0.600 & 0.553
    & 0.000 & 0.300 & 0.314
    & 15.29 & 0.426 & 0.628
    & 0.060 & 0.217 & 0.237 \\
    \rowcolor{orange!8}
    \cellcolor{white} & \cmark & \cmark
    & \textbf{14.23} & \textbf{0.426} & \textbf{0.685}
    & \textbf{0.017} & \textbf{0.097} & \textbf{0.107}
    & \textbf{8.57} & \textbf{0.197} & \textbf{0.754}
    & \textbf{0.220} & \textbf{0.077} & \textbf{0.071} \\

    \bottomrule
    \end{tabular}%
    }
\end{table*}

We conduct ablation studies on the two temporal designs: the intrinsic temporal modeling (ITM) and the surrogate temporal-attack loss (STA). Results are shown in \cref{tab:ablation}.

\textbf{Tuning-based pipeline.}
Removing the intrinsic consistency design lifts the constraint on the temporal shape of $\delta_y$, so on the original protected videos this variant achieves the strongest protection (PSNR $6.47$ vs. ours $12.17$, ISM1 $0.010$ vs. ours $0.160$). However, once any temporal filter is applied, the protection collapses, with ISM1 rising to $0.224$, $0.313$, and $0.300$ under the three filters, far higher than ours ($0.136$, $0.111$, $0.097$). Removing the surrogate loss also has an effect: protection on the original videos is stronger than ours, but under filters our full method consistently wins. The two ablations together show that each design trades a small amount of protection strength for a substantial gain in robustness.

\textbf{Reference-based pipeline.}
The reference-based pipeline shows a different pattern: our full method is the strongest on both the original protected videos and under filters. The reason is that the reference-based pipeline only consumes a single frame extracted from the protected video, so what matters is per-frame attack strength. Our temporal-consistency designs spread the perturbation evenly along time, so any extracted frame carries a strong attack. Without intrinsic consistency, the perturbation concentrates on a few frames and the extracted one may not be among them; this hurts even the original protection (PSNR $14.08$ vs. ours $8.00$, ISM1 $0.415$ vs. ours $0.275$) and degrades further under filters. Without the surrogate loss, performance is close to ours under most settings, with the surrogate giving a clear advantage on the low-pass filter (ISM1 $0.077$ vs. $0.093$). Consequently, our temporal-consistency designs add robustness without sacrificing per-frame attack strength.

\section{Conclusion}
\label{sec:conclusion}

In this work, we study video protection against unauthorized reference- and tuning-based customization, and identify three key temporal challenges: temporal compression, temporal overfitting, and temporal inconsistency. We propose \texttt{TC-UAP}, the first video-level protection method against both customization threats. By learning an identity-level multi-frame UAP from multiple videos of the same identity, \texttt{TC-UAP} generalizes to unseen videos and clips with different lengths, while intrinsic temporal modeling and an extrinsic surrogate temporal-attack loss improve robustness against temporal attacks. Extensive experiments across datasets and diffusion backbones show that \texttt{TC-UAP} achieves imperceptible, generalizable, and robust protection, taking a step toward securing video content against unauthorized customization.

\newpage
\bibliographystyle{plainnat}
\bibliography{main}

\newpage
\appendix
\appendixpage

\startcontents[sections]
\begingroup
\hypersetup{hidelinks}
\printcontents[sections]{l}{1}{\setcounter{tocdepth}{2}}
\endgroup
  
\clearpage
\section{Extended Methodology} \label{app:method}

\subsection{Analysis of Video Protection}
\label{app:rethinking}
\subsubsection{Failure of Image-Level Protection on Videos}
\label{app:rethinking-1}

\begin{figure}[h]
    \centering
    \includegraphics[width=0.8\textwidth]{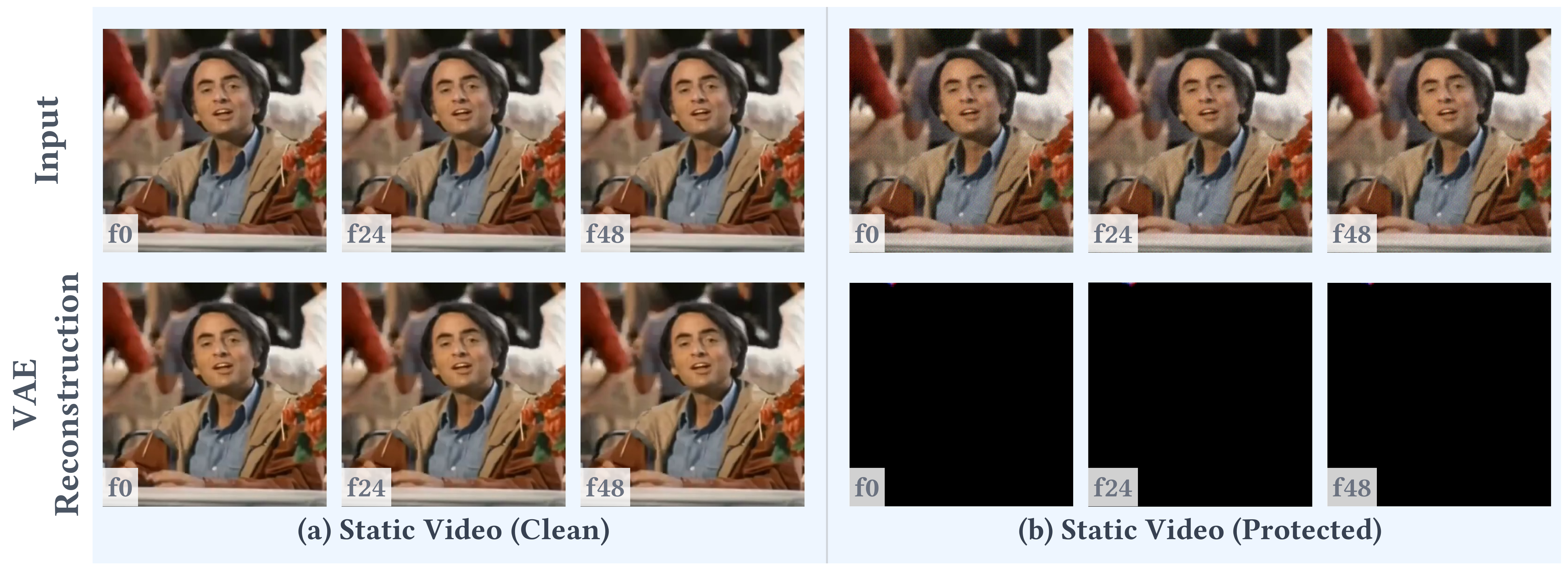}
    \caption{VAE reconstruction under the static video setting. (a) Clean static video. (b) Protected static video obtained by applying the same image protection to all frames.}
    \label{fig:static}
    \includegraphics[width=0.8\textwidth]{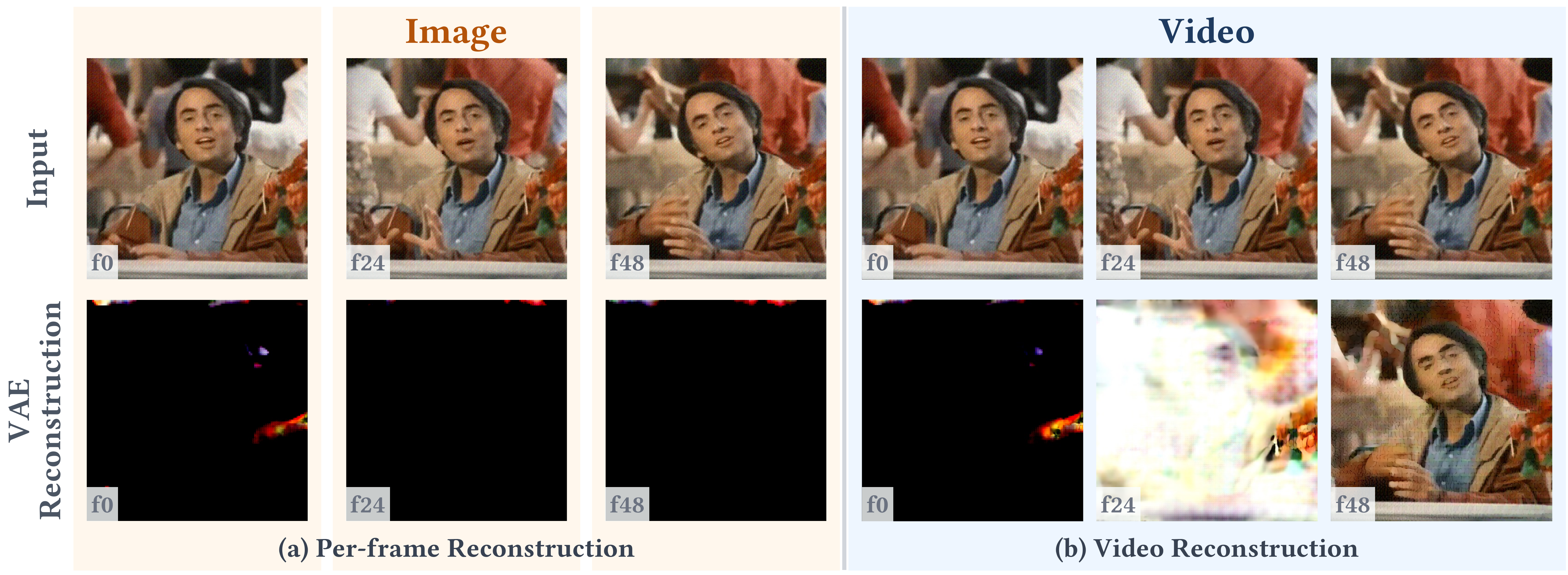}
    \caption{VAE reconstruction under the image-level UAP setting. (a) Protected frame reconstruction, where frames are extracted and reconstructed independently. (b) Video reconstruction.}
    \label{fig:1f_uap}
\end{figure}

\textbf{Static-content video and universal perturbations.} We further evaluate a static-video setting, where all frames are constructed by repeating the first protected frame (\textit{the same content and perturbations across frames}), and a universal adversarial perturbation (UAP) setting, where the same perturbation is applied to all frames (\textit{different content but same perturbations across frames}). As shown in \cref{fig:static} and \cref{fig:1f_uap}, the reconstructed video for \textit{static video} remains near-black but the reconstructed video for \textit{image-level UAP} still recovers in later frames just like the naive image-level perturbation. This suggests that the recovery observed in \cref{fig:obs1a}(d) \textit{does not come from the video VAE simply removing image-level perturbations from a single protected frame}. Instead, the recovery relies on temporal compression over different frames in the video, where information from previous frames can help restore the visual content. 

These additional results further support that frame-wise image perturbations cannot protect videos because they do not consider the temporal compression in 3D video VAEs.

\begin{figure}[h]
    \centering
    \includegraphics[width=0.8\textwidth]{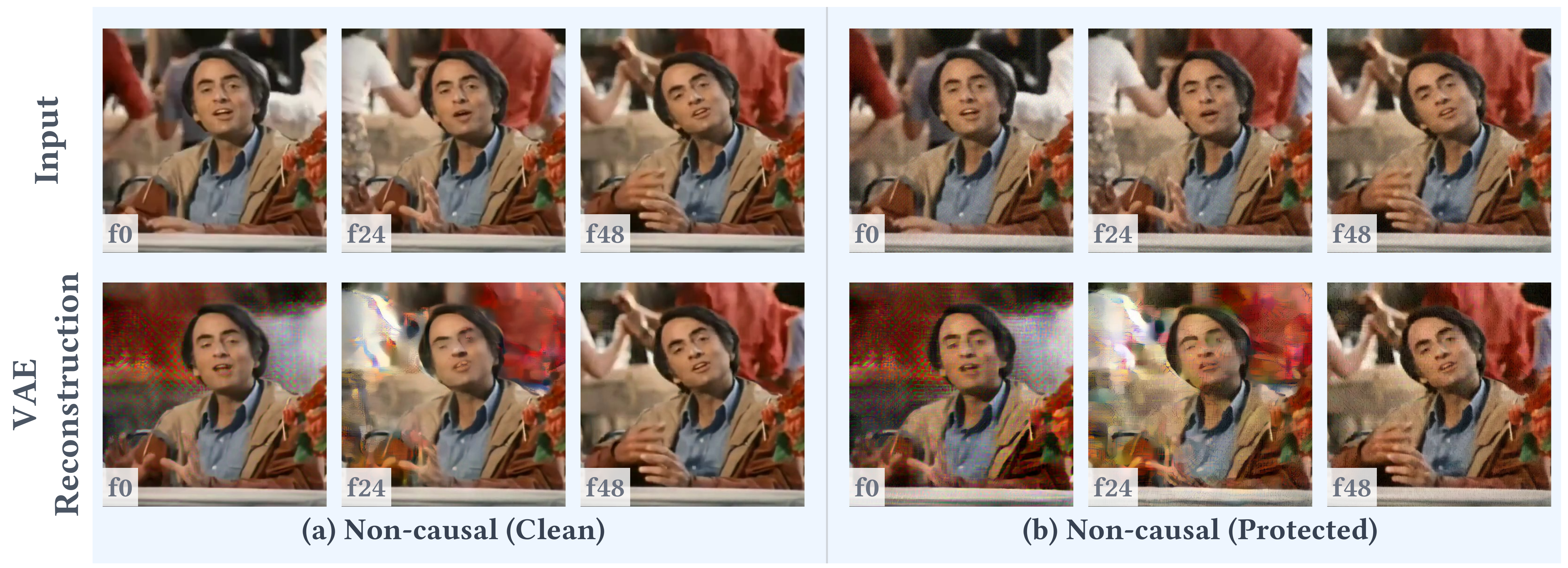}
    \caption{VAE reconstruction with a modified non-causal VAE encoder. (a) Clean video. (b) Per-frame protected video.}
    \label{fig:obs1_app}
\end{figure}

\textbf{Non-causal video VAE.} We provide additional results to further analyze the observation shown in \cref{fig:obs1a}(d). In the original LTX-2.3 VAE~\cite{hacohen2026ltx2efficientjointaudiovisual}, the encoder is causal along the temporal dimension, meaning that each latent position only compresses information from the current and \textit{previous frames} without accessing future frames. We modify the encoder into a \textbf{non-causal version} by allowing temporal convolution to also use \textit{future frames}. Since the VAE was originally trained with a causal encoder structure, this modification introduces some reconstruction mismatch. Therefore, we first evaluate clean video reconstruction under the non-causal encoder as a reference. As shown in \cref{fig:obs1_app}(a), the reconstructed clean video brings some distortion, but the main visual content remains recognizable.

We then use the same non-causal encoder to encode the per-frame protected video. As shown in \cref{fig:obs1_app}(b), the reconstruction recovers all content in the video, instead of only recovering in later frames as observed in \cref{fig:obs1a}(d). This indicates that the gradual recovery is related to the causal temporal structure of the video VAE. With non-causal encoding, earlier latent positions can also use information from later frames. Therefore, the temporal information used for reconstruction is no longer limited to previous frames, and the recovery is observed across all frames.

\subsubsection{Vulnerability to Temporal Attacks}
\label{app:rethinking-2}
We define the retention rate $\rho$ as the ratio of latent disruption after and before applying a temporal attack:
\begin{equation}
    \rho(\mathcal{T}, \delta_y) \;=\;
    \frac{\big\|\,\mathcal{E}(\mathcal{T}(V')) - \mathcal{E}(\mathcal{T}(V))\,\big\|_2^2}
         {\big\|\,\mathcal{E}(V') - \mathcal{E}(V)\,\big\|_2^2}.
    \label{eq:retention}
\end{equation}
A value of $\rho\!\to\!0$ indicates the perturbation is erased by $\mathcal{T}$, while $\rho\!\to\!1$ indicates the perturbation remains effective.

\textbf{Temporal attacks. } We evaluate seven temporal attacks used in \cref{fig:obs3}. Each attack is applied independently to every spatial-channel location $(c,h,w)$ along the temporal axis of $V \in \mathbb{R}^{T \times C \times H \times W}$. For clarity, the operations in \cref{tab:temporal-attacks} omit the $(c,h,w)$ subscripts, and indices outside $[0, T-1]$ are clamped to the nearest valid frame.

\begin{table}[h]
\centering
\small
\caption{Temporal attacks used in the retention analysis.}
\label{tab:temporal-attacks}
\begin{tabular}{p{0.22\linewidth} p{0.68\linewidth}}
\toprule
\textbf{Attack} & \textbf{Operation} \\
\midrule
\textsc{Avg-2}
& $\mathcal{T}(V)_t = 0.5\,V_t + 0.5\,V_{t+1}$ \\
\textsc{Skip}
& $\mathcal{T}(V)_t = 0.5\,V_{t-1} + 0.5\,V_{t+1}$ \\
\textsc{Avg-5}
& $\mathcal{T}(V)_t = \frac{1}{5}\sum_{i=-2}^{2} V_{t+i}$ \\
\textsc{Gauss}
& $\mathcal{T}(V)_t = \sum_{i=-r}^{r} g_i V_{t+i}$, where $g_i \propto \exp(-i^2/(2\sigma^2))$, $r=\lceil 3\sigma \rceil$, and $\sigma=2$ \\
\textsc{LPF}
& $\mathcal{T}(V)=\mathcal{F}_t^{-1}(M_k \odot \mathcal{F}_t(V))$, where $M_k(\omega)=1$ if $\min(\omega,T-\omega)<k$ and $0$ otherwise, with $k=8$ \\
\textsc{Down-$2\times$}
& $\mathcal{T}(V)=\mathrm{Up}_{2\times}(\mathrm{Down}_{2\times}(V))$ \\
\textsc{Drop-$30\%$}
& Randomly drop frames with $m_t\sim\mathrm{Bern}(1-p)$, $p=0.3$, and fill each dropped position with its nearest kept frame \\
\bottomrule
\end{tabular}
\end{table}

\subsection{Analysis of UAP}
\label{app:uap_cap}

\textbf{One-frame UAP does not provide sufficient temporal expressiveness for video protection.} One-frame UAP optimizes a single-frame perturbation and repeats it across all video frames. We analyze this design to test whether a one-frame UAP is sufficient for video protection.

\begin{figure}[h]
    \centering
    \includegraphics[width=\textwidth]{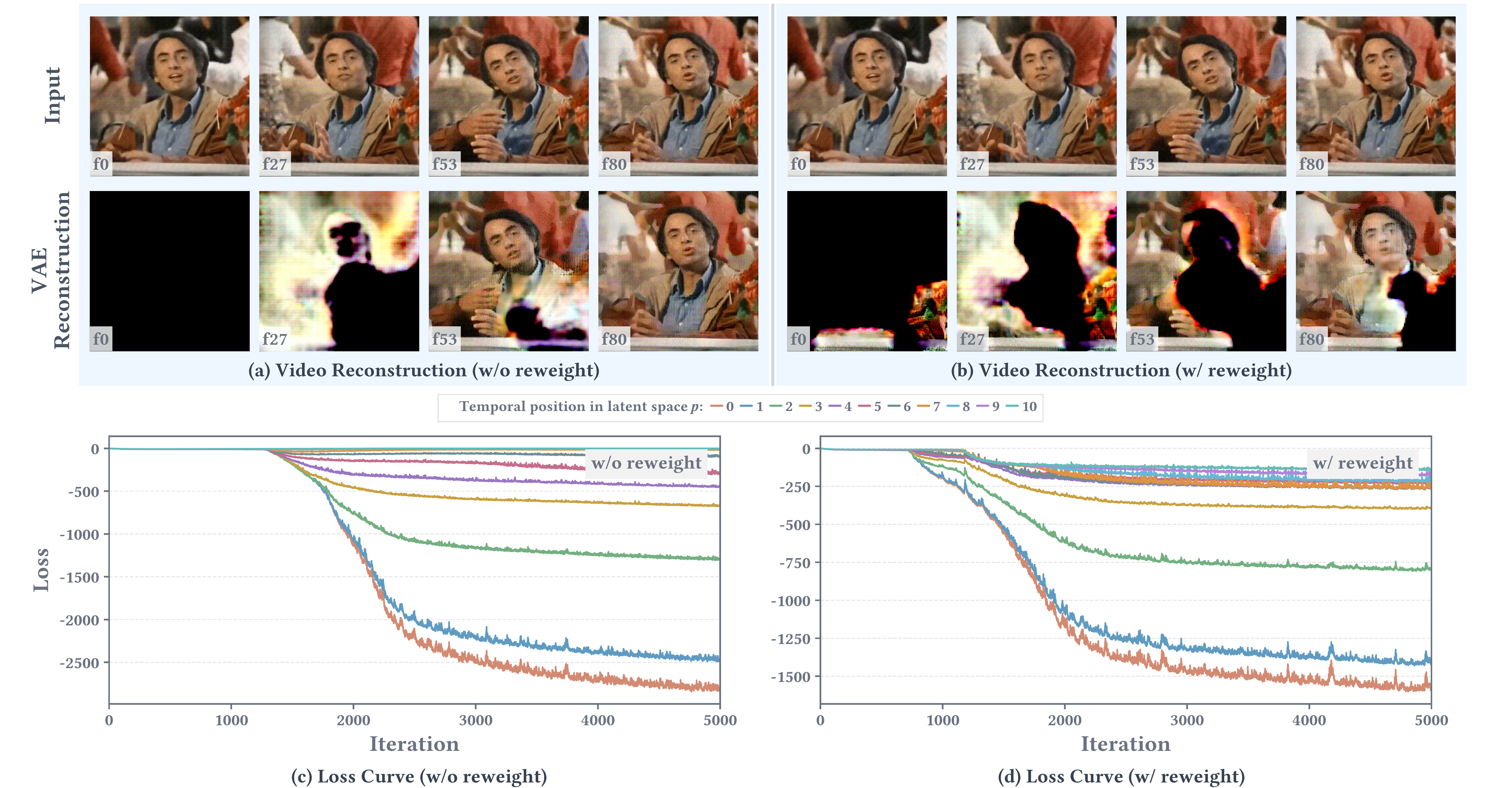}
    \caption{Analysis of one-frame UAP, where a single-frame UAP is shared across all frames and optimized against the video VAE. (a) Video reconstruction without reweight. (b) Video reconstruction with reweight. (c) Loss curves for different temporal latent positions without reweight. (d) Loss curves for different temporal latent positions with reweight.}
    \label{fig:uap_cap}
\end{figure}

We optimize the one-frame UAP with the same loss $\mathcal{L}_{\text{lat}}$ in Eq.~\eqref{eq:lat}, which maximizes the VAE latent distance between clean and protected videos. As shown in \cref{fig:uap_cap}(a) and (c), earlier temporal latent positions reach much lower losses, and the protection mainly appears in earlier frames, while later frames remain recognizable.

To examine whether this failure is caused by optimization imbalance, we \textit{reweight} the per-position losses by assigning each position a normalized weight inversely proportional to its loss magnitude. As shown in \cref{fig:uap_cap}(b) and (d), reweighting makes the loss more balanced across different positions and extends the disruption to later frames. However, the reconstructed video still contains recognizable content.

This indicates that the limitation of one-frame UAP is not only an optimization imbalance across temporal latent positions. It has limited temporal expressiveness to provide full-video protection.

\section{Extended Experimental Setup}
\label{app:exp-settings}

This section provides additional details for the experimental setup, including dataset construction, evaluation protocols, prompts, implementation settings, and baseline configurations.

\begin{figure}[htbp]
  \centering

  \begin{subfigure}{0.65\textwidth}
    \centering
    \includegraphics[width=\textwidth]{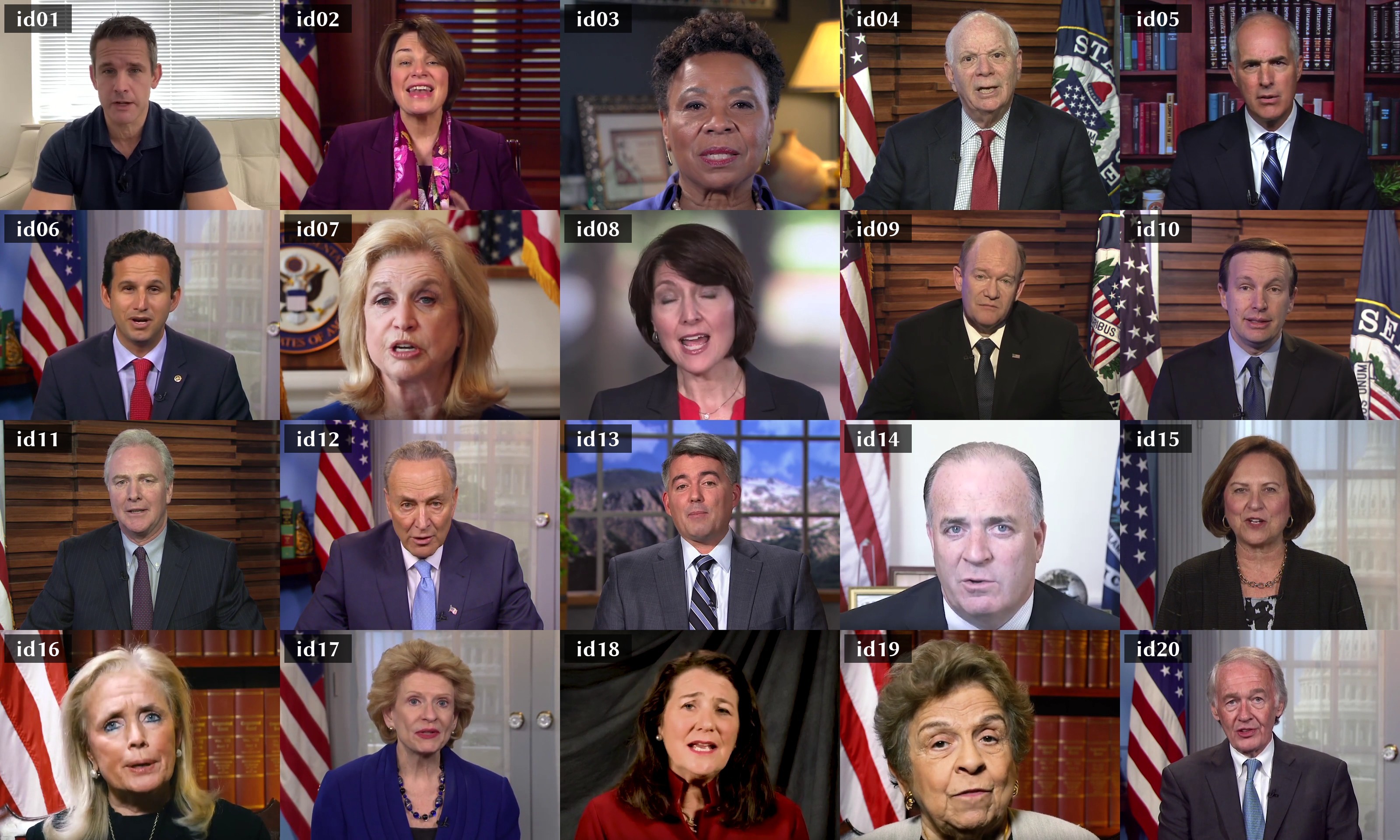}
    \caption{HDTF~\cite{zhang2021flow}}
    \label{fig:hdtf-id}
  \end{subfigure}

  \vspace{4pt}

  \begin{subfigure}{0.65\textwidth}
    \centering
    \includegraphics[width=\textwidth]{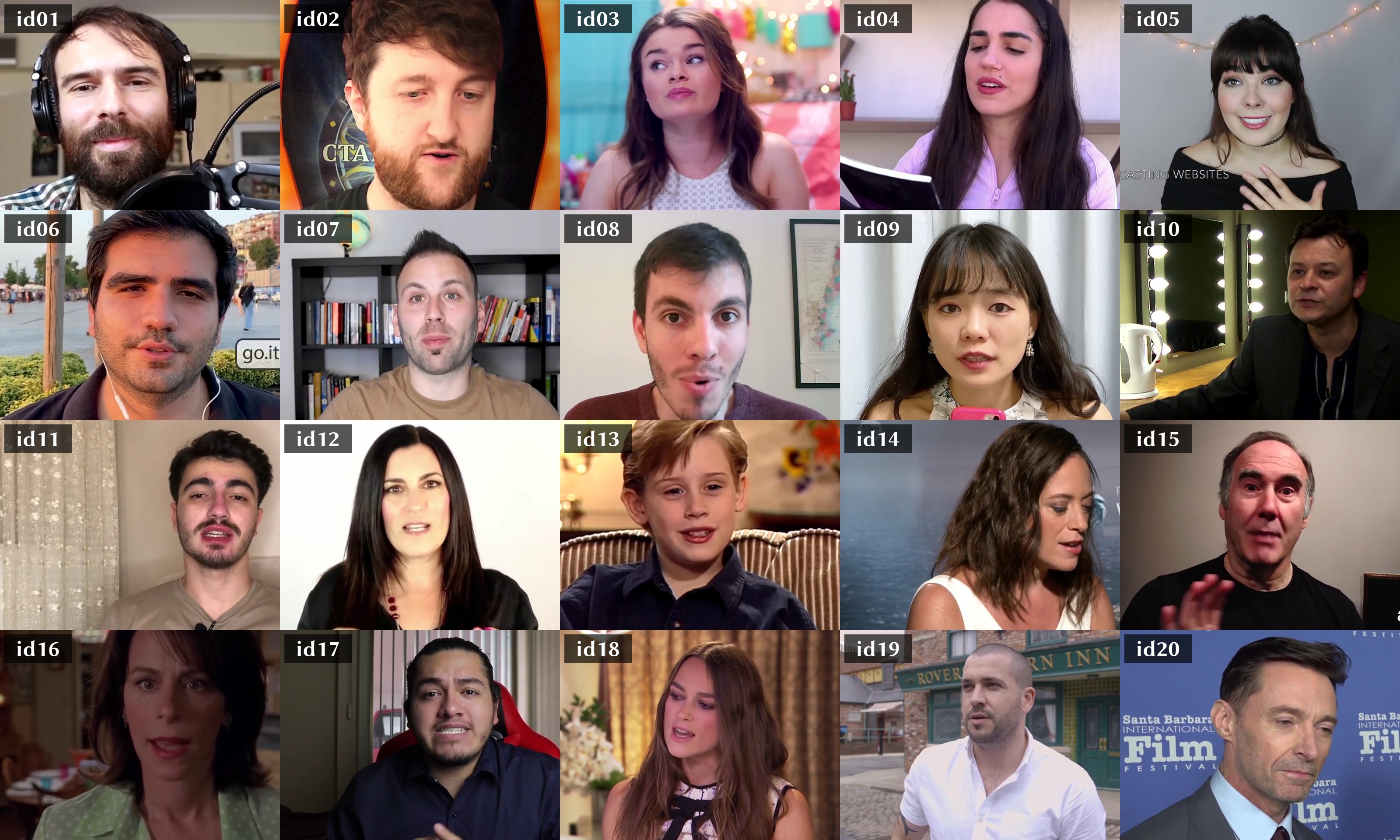}
    \caption{CelebV-HQ~\cite{zhu2022celebvhq}}
    \label{fig:celebvhq-id}
  \end{subfigure}

  \vspace{4pt}

  \begin{subfigure}{0.65\textwidth}
    \centering
    \includegraphics[width=\textwidth]{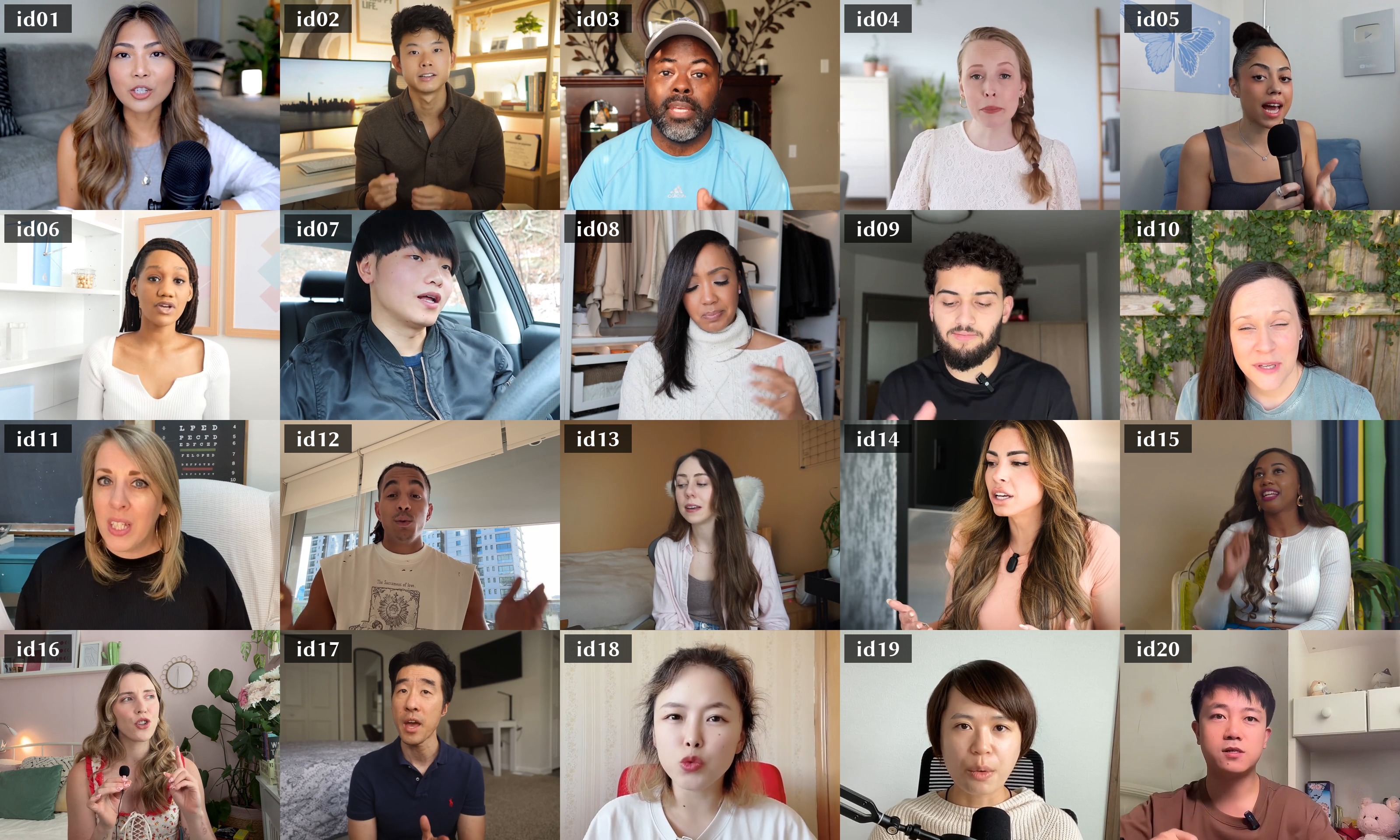}
    \caption{TalkVid~\cite{chen2025talkvid}}
    \label{fig:talkvid-id}
  \end{subfigure}

  \caption{Visualization of selected identities from the three datasets. One representative frame is shown for each identity.}
  \label{fig:dataset-vis}
\end{figure}

\subsection{Data Preparation}
\label{app:dataset-details}

To evaluate video protection against diffusion-based customization, we construct an evaluation set from three human-centered video datasets: HDTF~\cite{zhang2021flow}, CelebV-HQ~\cite{zhu2022celebvhq}, and TalkVid~\cite{chen2025talkvid}. These datasets are selected because they contain identity-centric videos, provide sufficient video samples per identity, and cover medium- to high-resolution human facial videos.

\textbf{HDTF. }
HDTF~\cite{zhang2021flow} is a large-scale, in-the-wild talking-face dataset collected from YouTube. It contains videos of 362 identities with a total duration of approximately 16 hours. The videos are provided at 720p or 1080p resolution and mainly contain frontal or near-frontal talking-head footage. Due to its relatively clean identity structure and high visual quality, HDTF is suitable for evaluating identity-level protection in talking-face videos.

\textbf{CelebV-HQ. }
CelebV-HQ~\cite{zhu2022celebvhq} is a large-scale high-quality celebrity video dataset designed for face video generation and facial attribute analysis. It contains 35,666 video clips from 15,653 identities. Compared with HDTF, CelebV-HQ contains more diverse appearance attributes, head poses, expressions, and visual conditions, making it useful for evaluating whether a protection method can generalize across more diverse facial videos.

\textbf{TalkVid. }
TalkVid~\cite{chen2025talkvid} is a large-scale talking-head video dataset containing 7,729 unique speakers and more than 1,244 hours of HD and 4K footage. It covers 15 languages and a broad age range. Compared with HDTF and CelebV-HQ, TalkVid provides more diverse talking-head videos in terms of speaker demographics, language, and recording scenarios.

\textbf{Identity selection and preprocessing. }
For each dataset, we select 20 identities. When metadata is available, such as in CelebV-HQ and TalkVid, we prioritize diversity in gender, race, and age group. We also prefer identities whose video clips cover multiple scenes, lighting conditions, and head poses. For datasets with limited metadata, we manually inspect candidate identities and select identities with clear facial visibility and sufficient video quality. Some visualization examples are shown in \cref{fig:dataset-vis}.

For each selected identity, we uniformly sample 30 video clips. Each clip is center-cropped around the face region and resized to $640 \times 480$. Each video clip contains 121 frames at 24 fps. We split the 30 clips of each identity into 15 training clips and 15 test clips. The training clips are used only for optimizing the identity-level universal adversarial perturbation (UAP), while the test clips are used for all downstream evaluations. After the UAP is trained, it is applied to the test clips to obtain the protected test set.

All datasets are used for non-commercial research purposes only and in accordance with their respective terms of use.
\begin{table}[htbp]
    \centering
    \caption{Datasets used in this work and their public sources.}
    \label{tab:dataset-licenses}
    \footnotesize
    \begin{tabular}{lll}
    \toprule
    \textbf{Dataset} & \textbf{Source URL} & \textbf{License / Terms of Use} \\
    \midrule
    HDTF~\cite{zhang2021flow}
        & \url{https://github.com/MRzzm/HDTF}
        & CC BY 4.0 \\
    CelebV-HQ~\cite{zhu2022celebvhq}
        & \url{https://github.com/CelebV-HQ/CelebV-HQ}
        & -- \\
    TalkVid~\cite{chen2025talkvid}
        & \url{https://github.com/FreedomIntelligence/TalkVid}
        & CC BY-NC 4.0 \\
    \bottomrule
    \end{tabular}
\end{table}

\subsection{Evaluation Metrics and Protocols}
\label{app:eval-details}

We evaluate our method from two perspectives: protection capability and imperceptibility. For protection capability, we report VAE reconstruction metrics and identity preservation metrics. For imperceptibility, we report invisibility metrics.

\textbf{VAE reconstruction metrics. }
Since our method targets the video VAE encoder, we need to evaluate how much the protected videos disrupt VAE reconstruction. Given a clean video $V$ and its protected version $V'$, we encode and decode the protected video using the VAE:
\begin{equation}
    \hat{V}' = \mathcal{D}(\mathcal{E}(V')).
\end{equation}
We then compare the reconstructed result $\hat{V}'$ against the clean input $V$ using PSNR, SSIM~\cite{wang2004image}, and LPIPS~\cite{zhang2018perceptual}.

PSNR measures pixel-level reconstruction fidelity:
\begin{equation}
    \mathrm{PSNR}(V, \hat{V}') =
    10 \log_{10} \left( \frac{\mathrm{MAX}^2}{\mathrm{MSE}(V, \hat{V}')} \right),
\end{equation}
where $\mathrm{MAX}$ is the maximum possible pixel value and $\mathrm{MSE}$ is the mean squared error between the clean and reconstructed videos. Lower PSNR indicates stronger VAE disruption.

SSIM measures structural similarity between the clean and reconstructed contents. Lower SSIM indicates that the reconstructed structure deviates more from the clean input. LPIPS measures perceptual distance using deep visual features; higher LPIPS indicates stronger perceptual distortion.

We report both video-level and image-level VAE reconstruction metrics. Video-level metrics are computed by reconstructing the protected video through the video VAE and are mainly associated with the tuning-based customization setting. Image-level metrics are computed on the conditioning frame used by the reference-based image-to-video pipeline.

\textbf{Identity preservation metrics. }
For generated videos, we also need to measure whether the identity in the generated video still matches the target identity. Hence, we report Face Detection Failure Rate (FDFR) and Identity Score Matching (ISM).

For each generated frame, we first use RetinaFace~\cite{Deng_2020_CVPR} to detect the face. If no face is detected, the frame is counted as a face detection failure. Given a generated video with $M$ frames, the Face Detection Failure Rate is defined as:
\begin{equation}
    \mathrm{FDFR} = \frac{M_{\mathrm{fail}}}{M},
\end{equation}
where $M_{\mathrm{fail}}$ is the number of frames where no face is detected. A higher FDFR indicates stronger protection, since the generated face becomes harder to detect.

For frames where a face is successfully detected, we compute face embeddings using two face recognition models: ArcFace~\cite{Deng_2019_CVPR} and CurricularFace~\cite{huang2020curricularface}. We report the corresponding identity similarity scores as ISM-ArcFace (ISM1) and ISM-Cur (ISM2). Given a generated frame embedding $f(\hat{I}_m)$ and a reference identity embedding $e_y$, the frame-level identity similarity is computed by cosine similarity:
\begin{equation}
    s_m = \frac{f(\hat{I}_m)^\top e_y}{\|f(\hat{I}_m)\|_2 \|e_y\|_2}.
\end{equation}
The video-level ISM is then averaged over all frames with successfully detected faces:
\begin{equation}
    \mathrm{ISM} = \frac{1}{|\Omega|} \sum_{m \in \Omega} s_m,
\end{equation}
where $\Omega$ denotes the set of generated frames with detected faces. Lower ISM indicates stronger identity protection.

For the tuning-based customization setting, the reference identity embedding $e_y$ is built from the clean video set of the same identity. Specifically, we uniformly sample 16 frames from the clean videos, detect the face in each frame, extract the face embeddings, and average them to obtain the identity reference. For the reference-based customization setting, the reference identity embedding is computed from the conditioning frame used by the image-to-video pipeline.

\textbf{Invisibility metrics. }
To evaluate whether the perturbation remains visually imperceptible, we measure the perceptual quality of the protected video $V'$ with respect to the clean video $V$ using Video Multi-Method Assessment Fusion (VMAF)~\cite{netflix2video}. VMAF is a full-reference video quality assessment metric that predicts human-perceived video quality by fusing multiple objective quality measurements, including detail loss, structural similarity, and motion-related features.

Given a clean video $V$ and its protected counterpart $V'$, we compute $\mathrm{VMAF}(V, V')$. 
A higher VMAF score indicates that the protected video is perceptually closer to the clean video, and therefore that the perturbation is less visible. In our evaluation, VMAF is used as the primary invisibility metric because it is designed for video-level perceptual quality assessment and accounts for temporal video characteristics beyond frame-wise image similarity.

\textbf{Evaluation prompts. }
For both tuning- and reference-based evaluations, we use the same set of five prompts to generate videos, as listed in \cref{tab:text-prompts}. The token ``\textcolor{gray}{\textit{p3r5on}}'' is used as the trigger token for the target identity in tuning-based customization.
\begin{table*}[htbp]
\centering
\caption{Text prompts used in the experiments.}
\label{tab:text-prompts}

\footnotesize
\setlength{\tabcolsep}{6pt}
\renewcommand{\arraystretch}{1.12}

\resizebox{0.95\textwidth}{!}{%
\begin{tabular}{cl}
\toprule
\textbf{ID} & \textbf{Prompt} \\
\midrule

$P_1$ & The \textcolor{gray}{\textit{p3r5on}} is licking an ice-cream cone, smiling between licks while facing the camera. \\

$P_2$ & The \textcolor{gray}{\textit{p3r5on}} takes a bite from a slice of pizza they are holding, chews slowly, and looks back at the camera. \\

$P_3$ & The \textcolor{gray}{\textit{p3r5on}} raises a coffee mug to their lips, takes a sip, sets it down, and continues facing the camera. \\

$P_4$ & The \textcolor{gray}{\textit{p3r5on}} opens a small notebook, glances at a page briefly, then closes it and looks at the camera. \\

$P_5$ & The \textcolor{gray}{\textit{p3r5on}} lifts a water bottle, takes a few sips, lowers it, and continues facing the camera. \\

\bottomrule
\end{tabular}%
}
\end{table*}
 
\textbf{Temporal attacks. }
For robustness evaluation, we use three unseen temporal attacks selected from the retention analysis in \cref{tab:temporal-attacks}: \textsc{Avg-5}, which denotes \textit{box filtering} with a temporal window size of 5; \textsc{Gauss}, which denotes \textit{Gaussian filtering} with $\sigma=2$; and \textsc{LPF}, which denotes \textit{low-pass filtering} with $k=8$.
These attacks are not used in the surrogate temporal-attack loss during optimization.

\subsection{Implementation Details}
\label{app:impl}

\textbf{UAP optimization. }
For each identity, we optimize an identity-level universal adversarial perturbation (UAP) on the 15 training clips. The UAP has a temporal length of 9 frames and is repeated along the temporal dimension to protect videos of arbitrary length. The perturbation is constrained by an $\ell_\infty$ budget of $\eta = 0.1$.

We optimize the UAP against the video VAE encoder. The $\lambda$ in \cref{eq:total-loss} is set to $1.0$. The learning rate is set to $4 \times 10^{-3}$, and the UAP is optimized for one epoch over the training clips. After each optimization step, the perturbation is projected back to the $\ell_\infty$ ball.

\textbf{Tuning-based customization. }
For tuning-based evaluation, we fine-tune LoRA adapters on the protected test videos of each identity. The training caption uses the trigger token ``\textcolor{gray}{\textit{p3r5on}}''. We use AdamW as the optimizer with learning rate $1 \times 10^{-4}$ and batch size 1. For LTX-2.3~\cite{hacohen2026ltx2efficientjointaudiovisual}, LoRA fine-tuning is performed for 1,500 steps. For Wan2.2-5B~\cite{wan2025wanopenadvancedlargescale}, LoRA fine-tuning is performed for 2,000 steps.

After fine-tuning, we generate videos using the evaluation prompts described in Appendix~\ref{app:eval-details}. The generated videos are evaluated using identity preservation metrics, including FDFR, ISM-ArcFace, and ISM-Cur.

\textbf{Reference-based customization. }
For reference-based evaluation, we use the official image-to-video pipelines of LTX-2.3 and Wan2.2-5B. Each generated video is conditioned on one frame extracted from a protected test video. The same evaluation prompts are used as in the tuning-based setting. Since the reference-based pipeline directly consumes a protected frame, we additionally report image-level VAE reconstruction metrics on the conditioning frame.

\textbf{Hardware. }
UAP optimization is conducted on a single NVIDIA RTX 4090 GPU (24G). LoRA fine-tuning and video generation are performed on a single NVIDIA RTX Pro 6000 GPU (96G).

\section{Additional Experimental Results}
\label{app:qualitative}

\begin{figure*}[h]
    \centering
    \includegraphics[width=\textwidth]{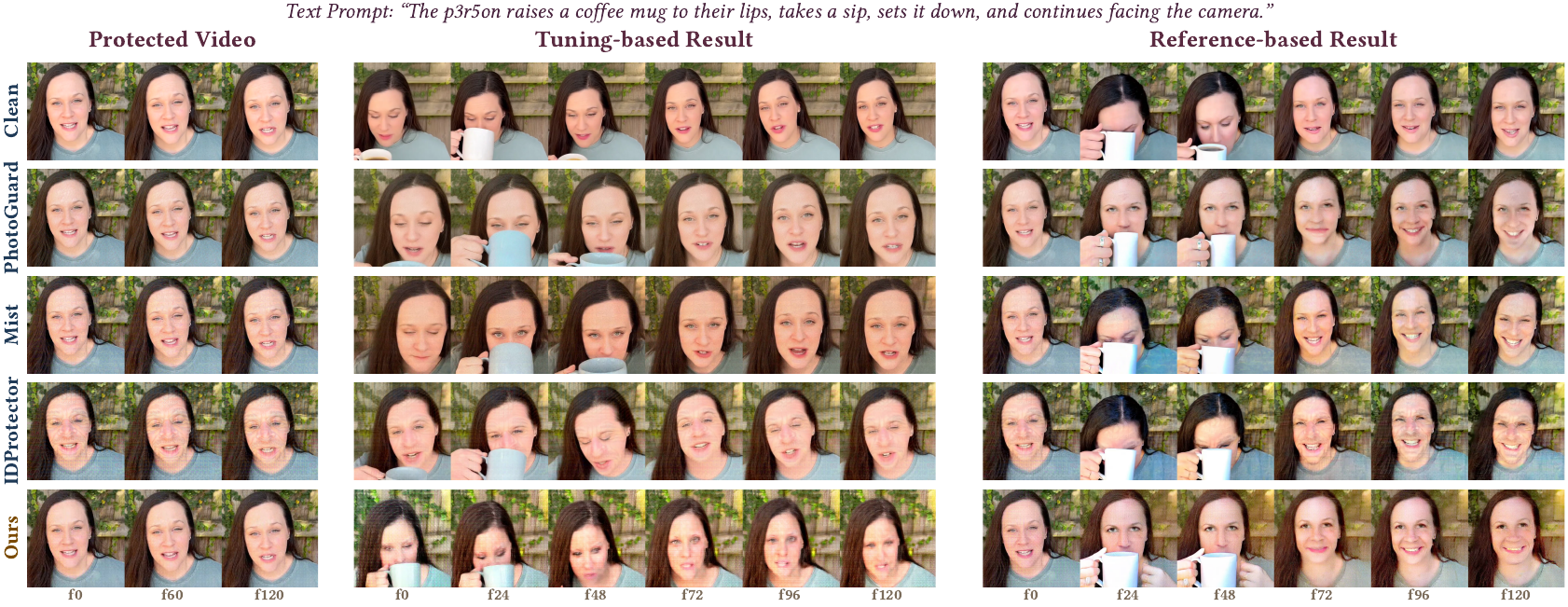}
    \caption{Qualitative comparison of customization results on the Wan2.2-5B model~\cite{wan2025wanopenadvancedlargescale}.}
    \label{fig:wan}
\end{figure*}

\begin{figure*}[h]
    \centering
    \begin{subfigure}[b]{\textwidth}
        \includegraphics[width=\linewidth]{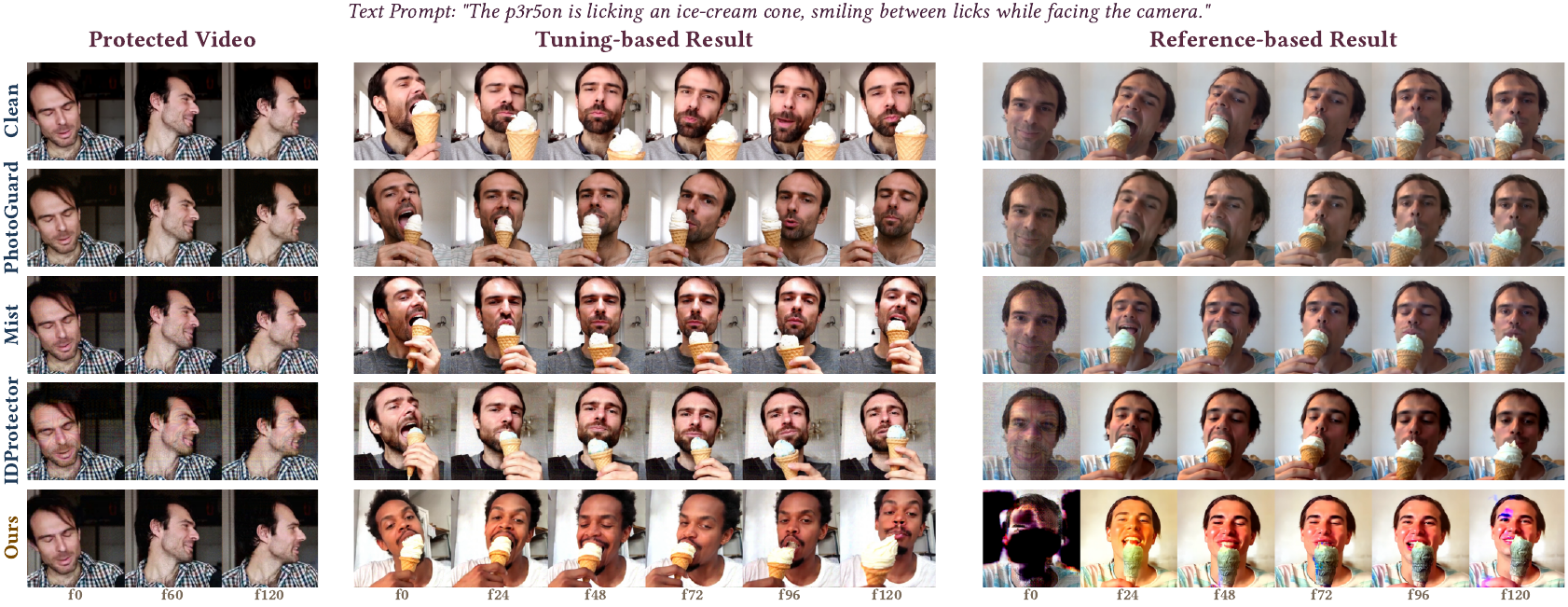}
        \caption{}
        \label{fig:CelebV-a}
    \end{subfigure}

    \vspace{4pt}

    \begin{subfigure}[b]{\textwidth}
        \includegraphics[width=\linewidth]{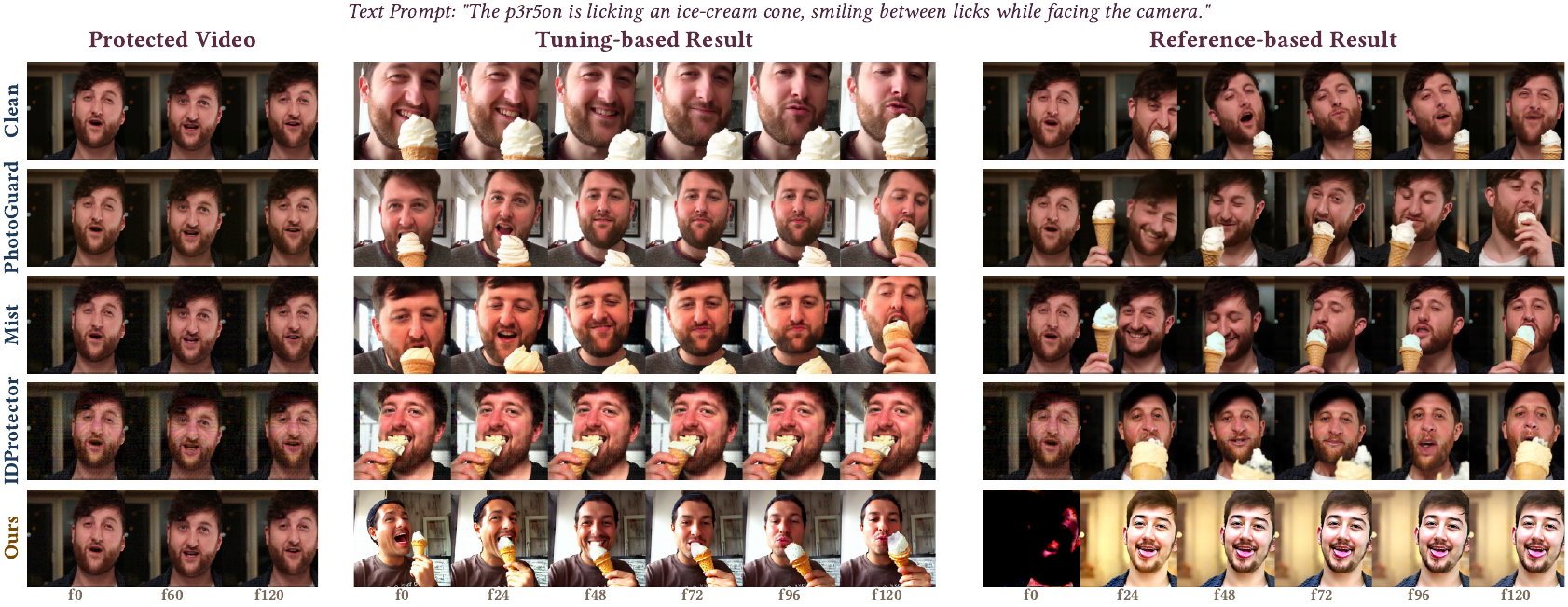}
        \caption{}
        \label{fig:CelebV-b}
    \end{subfigure}

    \vspace{4pt}

    \begin{subfigure}[b]{\textwidth}
        \includegraphics[width=\linewidth]{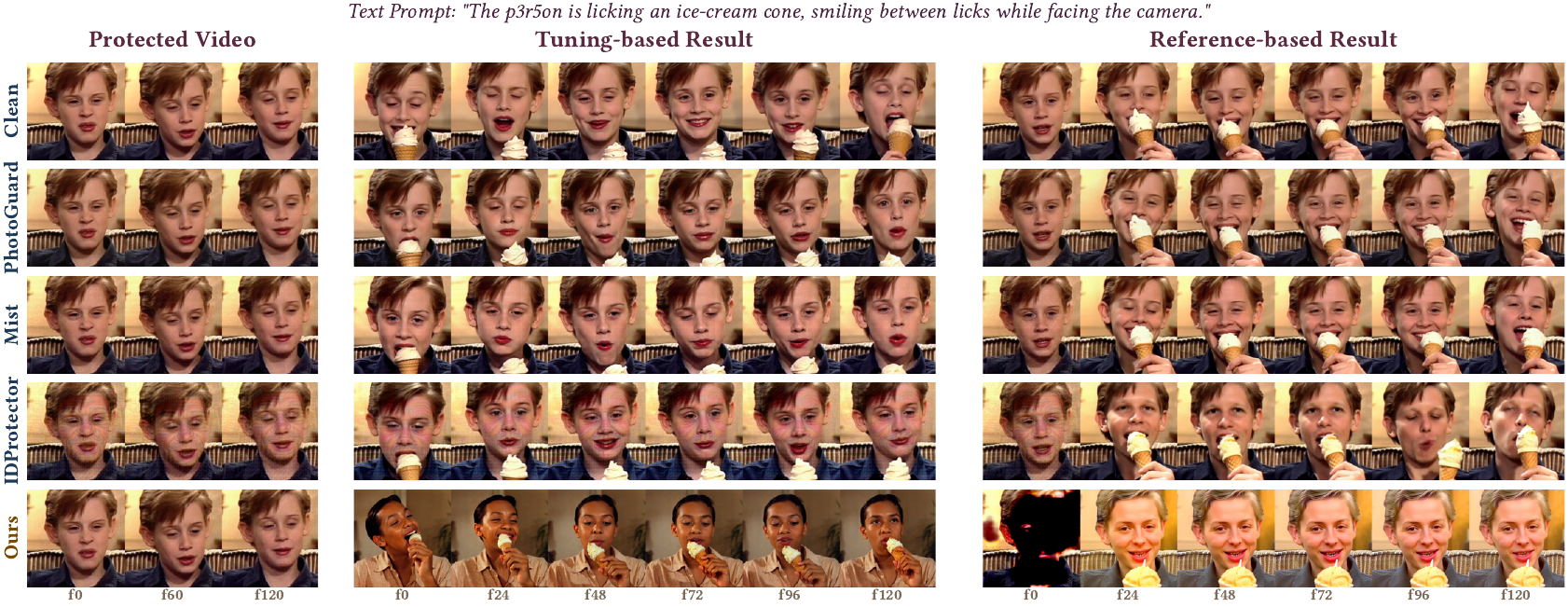}
        \caption{}
        \label{fig:CelebV-d}
    \end{subfigure}
    \caption{Qualitative comparison on CelebV-HQ Dataset.}
    \label{fig:CelebV}
\end{figure*}

\cref{fig:wan} and \cref{fig:CelebV} provide additional qualitative comparisons on LTX-2.3~\cite{hacohen2026ltx2efficientjointaudiovisual} and Wan2.2-5B~\cite{wan2025wanopenadvancedlargescale}. These results further demonstrate that \texttt{TC-UAP} provides effective protection under both tuning- and reference-based customization pipelines.

\newpage

\newpage
\begin{figure*}[h]
    \centering
    \begin{subfigure}[b]{\textwidth}
        \includegraphics[width=\linewidth]{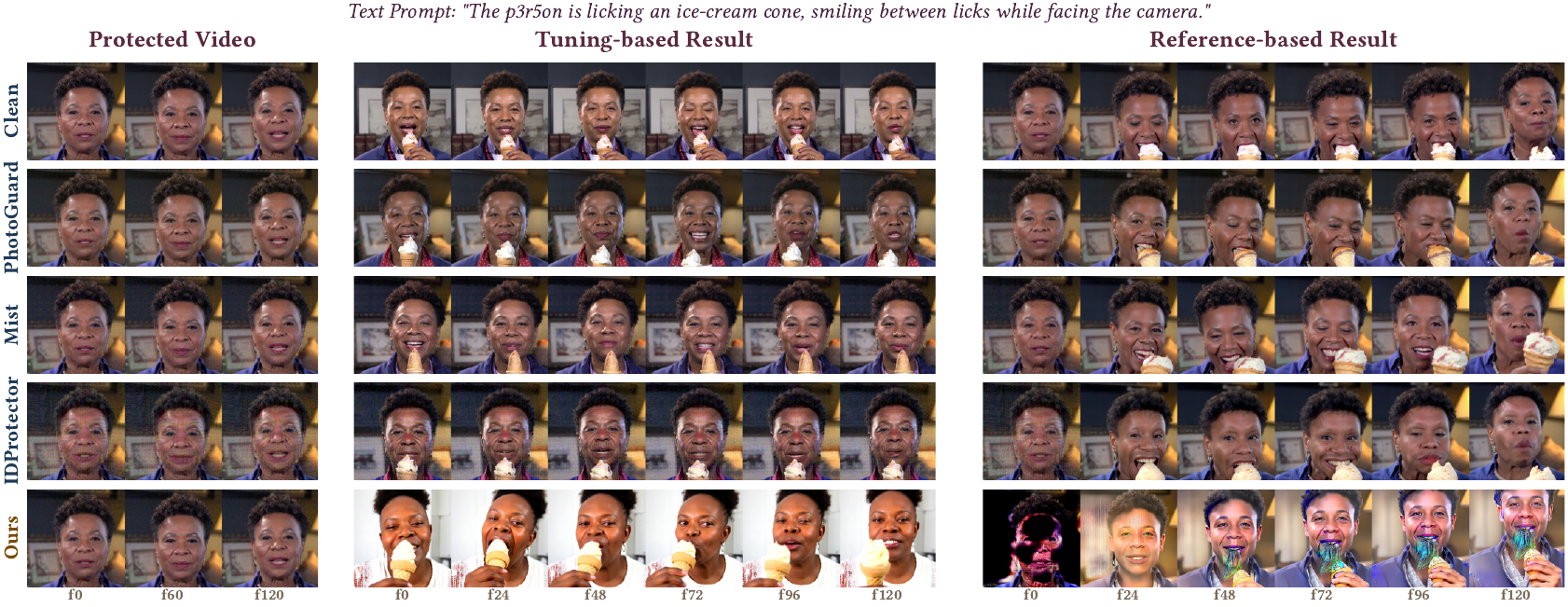}
        \caption{}
        \label{fig:HDTF-a}
    \end{subfigure}

    \vspace{4pt}

    \begin{subfigure}[b]{\textwidth}
        \includegraphics[width=\linewidth]{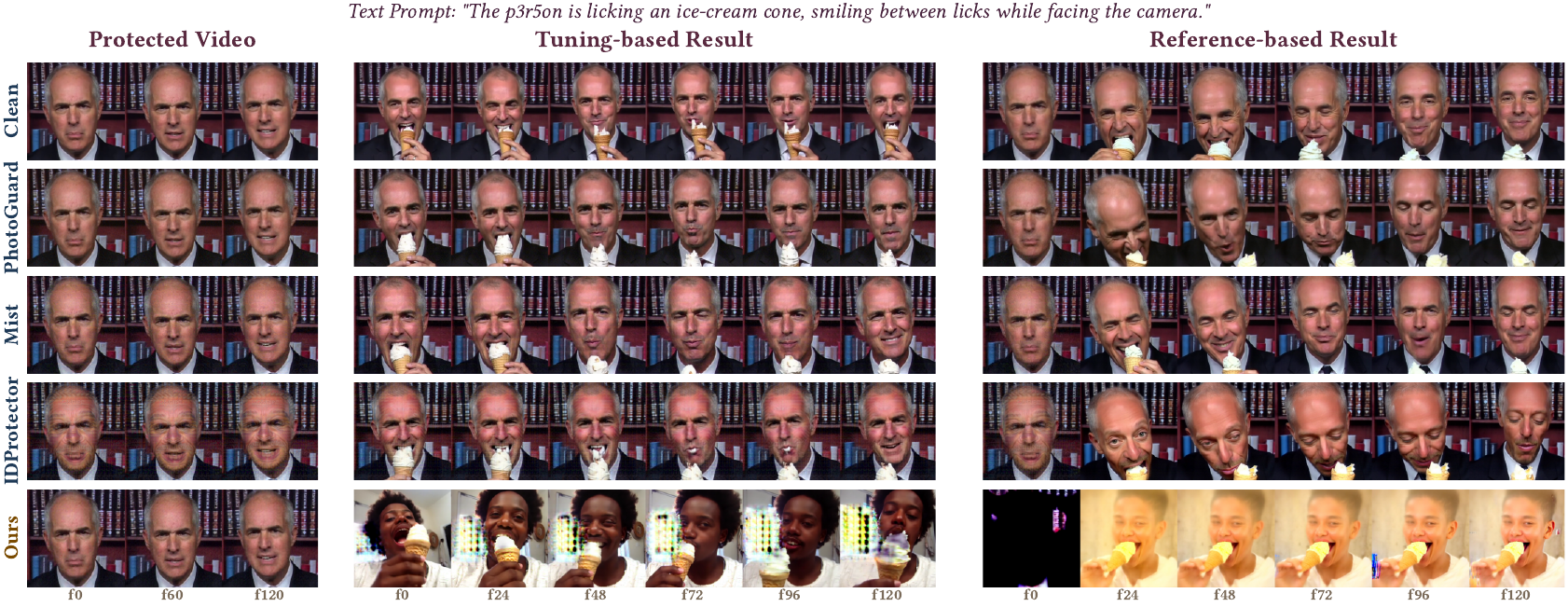}
        \caption{}
        \label{fig:HDTF-b}
    \end{subfigure}

    \vspace{4pt}

    \begin{subfigure}[b]{\textwidth}
        \includegraphics[width=\linewidth]{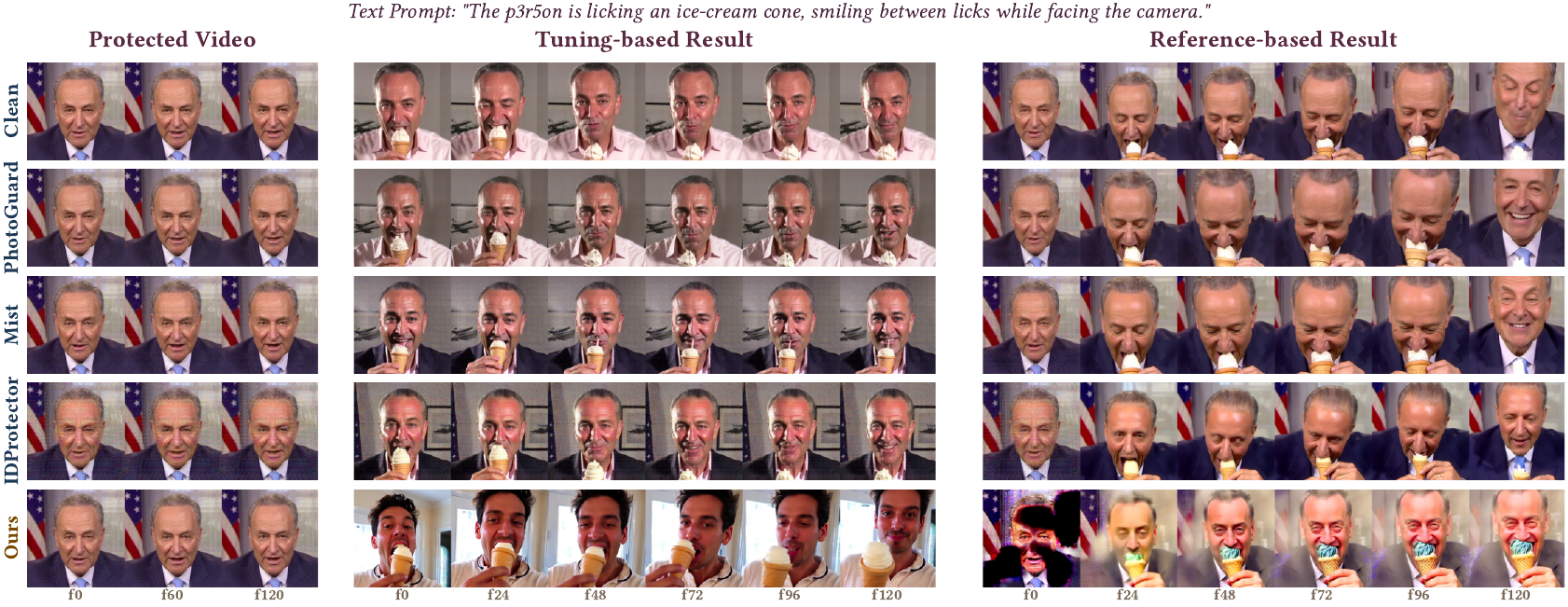}
        \caption{}
        \label{fig:HDTF-d}
    \end{subfigure}
    \caption{Qualitative comparison on HDTF Dataset.}
    \label{fig:HDTF}
\end{figure*}

\newpage
\begin{figure*}[h]
    \centering
    \begin{subfigure}[b]{\textwidth}
        \includegraphics[width=\linewidth]{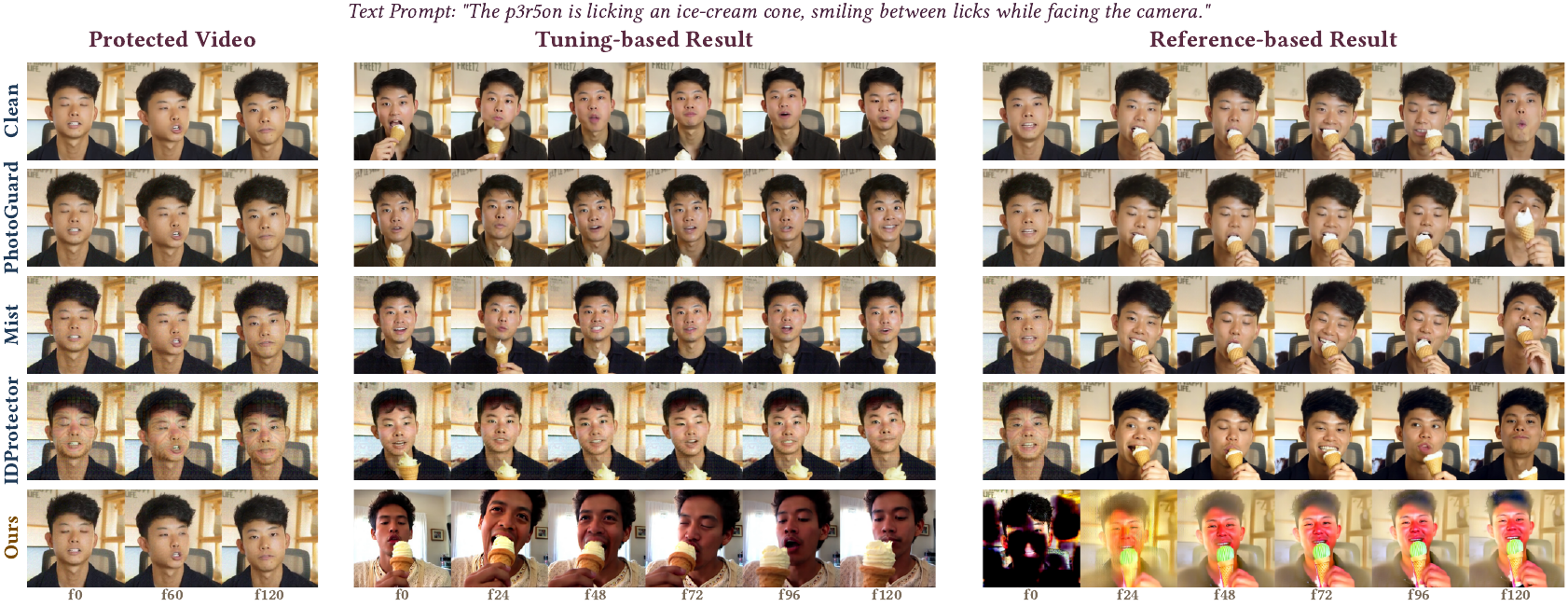}
        \caption{}
        \label{fig:TalkVid-a}
    \end{subfigure}

    \vspace{4pt}

    \begin{subfigure}[b]{\textwidth}
        \includegraphics[width=\linewidth]{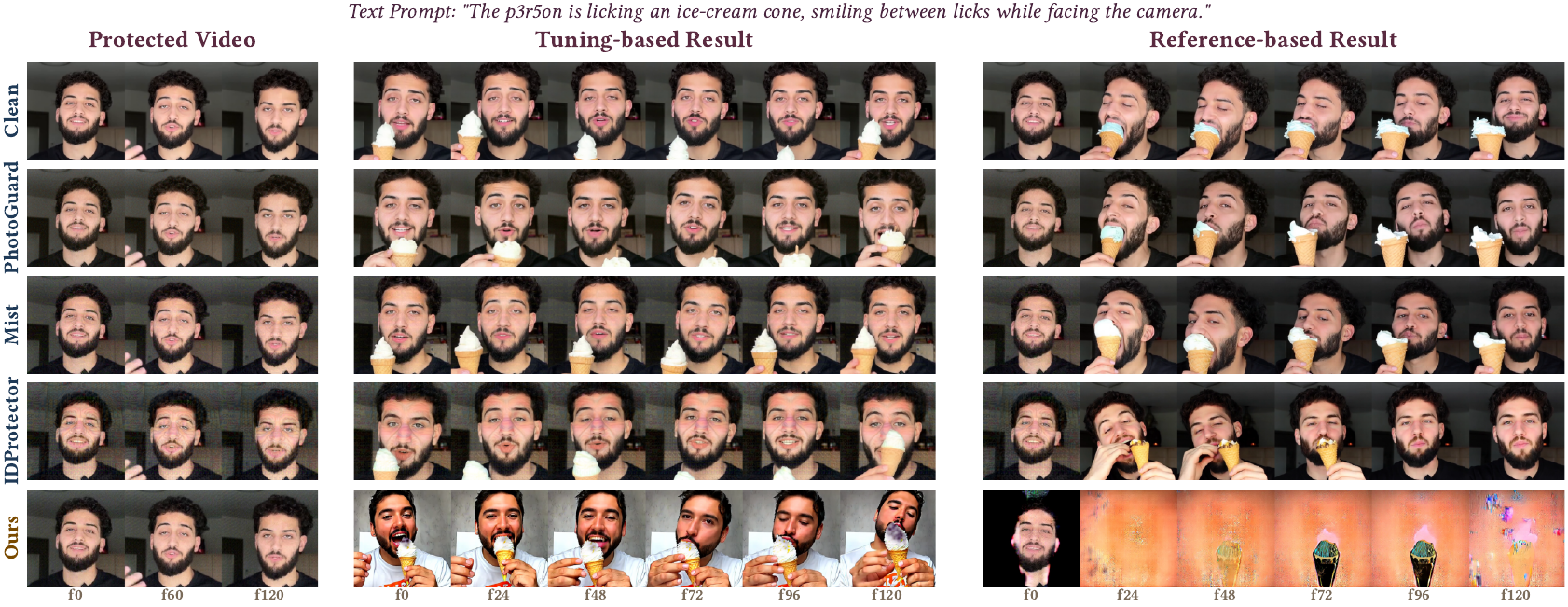}
        \caption{}
        \label{fig:TalkVid-b}
    \end{subfigure}

    \vspace{4pt}

    \begin{subfigure}[b]{\textwidth}
        \includegraphics[width=\linewidth]{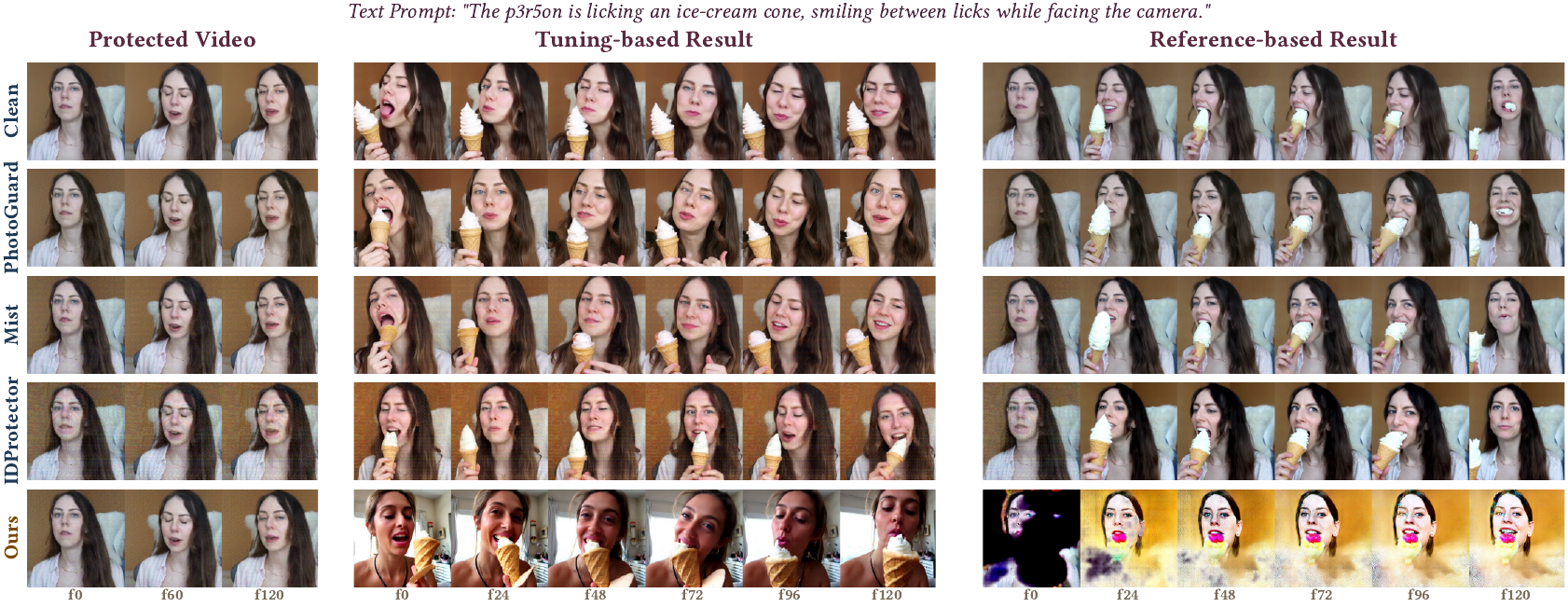}
        \caption{}
        \label{fig:TalkVid-d}
    \end{subfigure}
    \caption{Qualitative comparison on TalkVid Dataset.}
    \label{fig:TalkVid}
\end{figure*}

\clearpage
\section{Limitations and Broader Impacts}

\textbf{Limitations. } 
While \texttt{TC-UAP} provides strong protection against unauthorized identity customization in diffusion-based video generation, it primarily targets the visual VAE latent space. However, a person's identity in video content also includes acoustic characteristics (e.g., voice and speech patterns). As a result, our current method does not prevent attackers from extracting or cloning the audio track of a protected video. Developing joint audio-visual protection that disrupts both the visual VAE and the audio tokenizer or encoder is a promising direction for future work.

\textbf{Broader impacts. }
Personal videos posted online can be collected without the subject's authorization and used to fine-tune diffusion models or to condition image-to-video pipelines, producing customized content the subject never agreed to. \texttt{TC-UAP} gives individual users a practical tool to limit this: applying an imperceptible perturbation to a video before release leaves it visually unchanged for human viewers but renders it ineffective as training material for tuning-based customization or as a reference for I2V generation. The default thus shifts from ``any uploaded video can be reused for identity-preserving generation'' to ``the uploader decides whether their identity can be replicated downstream.'' Content creators, journalists, and ordinary users can then share footage on public platforms while retaining control over how their face and likeness are reused.

At the policy level, opt-out clauses in current data-use regulations depend on data collectors voluntarily honoring them and are easily bypassed by automated scraping. \texttt{TC-UAP} complements such policies with a technical enforcement layer that does not require collector cooperation: even when a protected video is scraped, it provides no usable signal for downstream customization. This raises the cost of unauthorized identity replication and reduces the effective supply of clean training material for non-consensual deepfakes. We note two limitations of this perspective. First, protection applies only to videos released after the tool is adopted; previously released material cannot be retroactively shielded. Second, as with any defensive technique, \texttt{TC-UAP} may be evaded by future attackers with stronger temporal models or by collectors who obtain data from offline sources, so it should be viewed as one component of a broader set of privacy measures rather than a complete solution.

\end{document}